%% file: main.tex
\RequirePackage{snapshot}
\documentclass{article}

\usepackage{mycustom}
\usepackage{microtype}
\usepackage{paralist} % compactenunm, compactitem

\usepackage[accepted]{icml2018}

\newcommand{\ourtitle}{Fixing a Broken ELBO}
\icmltitlerunning{Fixing a Broken ELBO}

\newcites{supref}{Supplementary References}
\begin{document}

\twocolumn[
	\icmltitle{\ourtitle}




\icmlsetsymbol{dm}{*}

\begin{icmlauthorlist}
	\icmlauthor{Alexander A.~Alemi}{goo}
	\icmlauthor{Ben Poole}{stan,dm}
	\icmlauthor{Ian Fischer}{goo}
	\icmlauthor{Joshua V.~Dillon}{goo}
	\icmlauthor{Rif A.~Saurous}{goo}
	\icmlauthor{Kevin Murphy}{goo}
\end{icmlauthorlist}

\icmlaffiliation{goo}{Google AI}
\icmlaffiliation{stan}{Stanford University}

\icmlcorrespondingauthor{Alexander A.~Alemi}{alemi@google.com}

\icmlkeywords{Machine Learning, ICML, Unsupervised Learning, Variational Autoencoder, VAE, Generative Models}

\vskip 0.3in
]



\printAffiliationsAndNotice{\textsuperscript{*}Work done during an internship at DeepMind. }  % leave blank if no need to mention equal contribution

\newcommand{\eat}[1]{}

	\InputIfFileExists{abstract}{}{\typeout{No file abstract.}}%

	\InputIfFileExists{intro}{}{\typeout{No file intro.}}%
	
%
	\InputIfFileExists{framework}{}{\typeout{No file framework.}}%

	\InputIfFileExists{related}{}{\typeout{No file related.}}%

	\InputIfFileExists{experiments}{}{\typeout{No file experiments.}}%

	\InputIfFileExists{concl}{}{\typeout{No file concl.}}%

\bibliographystyle{icml2018}
\bibliography{bib}

\clearpage
\onecolumn
\appendix

\begin{center}
	\textbf{\large Supplemental Materials: \ourtitle}
\end{center}

	\InputIfFileExists{mnistextra}{}{\typeout{No file mnistextra.}}%

	\InputIfFileExists{omniextra}{}{\typeout{No file omniextra.}}%

	\InputIfFileExists{generative}{}{\typeout{No file generative.}}%

	\InputIfFileExists{proofs}{}{\typeout{No file proofs.}}%
	
%\inputfile{relatedextra}
%
	\InputIfFileExists{toymodelextra}{}{\typeout{No file toymodelextra.}}%

	\InputIfFileExists{mnistdetails}{}{\typeout{No file mnistdetails.}}%

\bibliographystylesupref{icml2018}
\bibliographysupref{bib}

\end{document}

%% file: abstract.tex
\begin{abstract}

Recent work in unsupervised representation learning has focused on 
learning deep directed latent-variable models. 
Fitting these models by maximizing the
marginal likelihood or evidence is typically intractable, thus a common
approximation is to maximize the evidence lower bound (ELBO) instead.
However, maximum likelihood training (whether exact or
approximate)
does not necessarily result in a good latent
representation, as we demonstrate both theoretically and
empirically. 
In particular, we derive variational lower and
upper bounds on the mutual information between the input and the latent
variable,
and use these bounds to derive a rate-distortion curve that
characterizes the tradeoff between compression and
reconstruction accuracy. Using this framework, we
demonstrate that there is a family of models with identical
ELBO, but different quantitative and qualitative characteristics.
Our framework also suggests a simple new method to ensure that
latent variable
models with powerful stochastic decoders do not ignore their
latent code.

\end{abstract}

%% file: intro.tex
\section{Introduction}
\label{sec:intro}

Learning a ``useful'' representation of data in an unsupervised way is one of
the ``holy grails'' of current machine learning research.  A common approach to
this problem is to fit a latent variable model of the form
$p(x,z|\theta)=p(z|\theta) p(x|z,\theta)$ to the data, where $x$ are the
observed variables, $z$ are the hidden variables, and $\theta$ are the
parameters.  We usually fit such models by minimizing $L(\theta) =
\KL{\qemp(x)}{p(x|\theta)}$, which is equivalent to maximum likelihood
training.  If this is intractable, we may instead maximize a lower bound on
this quantity, such as the evidence lower bound (ELBO), as is done when fitting
variational autoencoder (VAE) models \citep{vae,Rezende14}.  Alternatively, we
can consider other divergence measures, such as the reverse KL, $L(\theta) =
\KL{p(x|\theta)}{\qemp(x)}$, as is done when fitting certain kinds of
generative adversarial networks (GANs).  However, the fundamental problem is
that these loss functions only depend on $p(x|\theta)$, and not on
$p(x,z|\theta)$. Thus they do not measure the quality of the representation at
all, as discussed  in \citep{huszar,mutualae}.  In particular, if we have a
powerful stochastic decoder $p(x|z,\theta)$, such as an RNN or PixelCNN, a VAE
can easily ignore $z$ and still obtain high marginal likelihood $p(x|\theta)$,
as noticed in \citep{bowman2015generating,vlae}.  Thus obtaining a good ELBO
(and more generally, a good marginal likelihood) is not enough for good
representation learning.

In this paper, we argue that a better way to assess the value of representation
learning is to measure the mutual information $I$ between the observed $X$ and
the latent $Z$.  In general, this quantity is intractable to compute, but we
can derive tractable variational lower and upper bounds on it.  By varying $I$,
we can tradeoff between how much the data has been compressed vs how much
information we retain.  This can be expressed using the rate-distortion or $RD$
curve from information theory, as we explain in \cref{sec:framework}.  This
framework provides a solution to the problem of powerful decoders ignoring the
latent variable which is simpler than the architectural constraints of
\citep{vlae}, and more general than the ``KL annealing'' approach of
\citep{bowman2015generating}.  This framework also generalizes  the $\beta$-VAE
approach used in \citep{betavae,vib}.  \eat{ In particular, although changing
$\beta$ does allow us to move along the $RD$ curve, it cannot target arbitrary
values of the mutual information, unlike our more general approach.  }

In addition to our unifying theoretical framework, we empirically study the
performance of a variety of different VAE models --- with both ``simple'' and
``complex'' encoders, decoders, and priors --- on several simple image datasets
in terms of the $RD$ curve. We show that VAEs with powerful autoregressive
decoders can be trained to not ignore their latent code by targeting certain
points on this curve.  We also show how it is possible to recover the ``true
generative process'' (up to reparameterization) of a simple model on a
synthetic dataset with no prior knowledge except for the true value of the
mutual information $I$ (derived from the true generative model).  We believe
that information constraints provide an interesting alternative way to
regularize the learning of latent variable models.

%% file: framework.tex
%\section{Informative Representational Learning}
\section{Information-theoretic framework}
\label{sec:framework}

In this section, we outline our information-theoretic view of
\emph{unsupervised} representation learning.  Although many of these ideas have
been studied in prior work (see \cref{sec:related}), we provide a unique
synthesis of this material into a single coherent, computationally tractable
framework.  In \cref{sec:experiments}, we show how to use this framework to
study the properties of various recently-proposed VAE model variants.

\input{representation}
\input{phase}
\input{betavae}

%% file: representation.tex
\paragraph{Unsupervised Representation Learning}

We will convert each observed data vector $x$ into a latent representation $z$
using any stochastic encoder $\qe(z|x)$ of our choosing. This then induces the
joint distribution $\qprep(x,z) = \qp(x) \qe(z|x)$ and the corresponding
marginal posterior $\qprep(z) = \int dx\, \qp(x) \qe(z|x)$ (the ``aggregated
posterior'' in \citet{makhzani2015adversarial, vampprior}) and conditional
$\qprep(x|z) = \qprep(x,z)/\qprep(z)$.

Having defined a joint density, a symmetric, non-negative,
reparameterization-independent measure of how much information one random
variable contains about the other is given by the mutual information:
\begin{equation}
	\Irep(X; Z) = \iint dx\, dz\, \qprep(x,z) \log \frac{\qprep(x,z)}{\qp(x)\qprep(z)}.
	\label{eqn:Irep}
\end{equation}
(We use the notation $\Irep$ to emphasize the dependence on our choice of
encoder. See \cref{sec:Igen} for other definitions of mutual information.)
There are two natural limits the mutual information can take. In one extreme,
$X$ and $Z$ are independent random variables, so the mutual information
vanishes: our representation contains no information about the data whatsoever.
In the other extreme, our encoding might just be an identity map, in which
$Z=X$ and the mutual information becomes the entropy in the data $H(X)$.  While
in this case our representation contains all information present in the
original data, we arguably have not done anything meaningful with the data.  As
such, we are interested in learning representations with some fixed mutual
information, in the hope that the information $Z$ contains about $X$ is in some
ways the most salient or useful information.

Equation~\ref{eqn:Irep} is hard to compute, since we do not have access to the
true data density $\qp(x)$, and computing the marginal $\qprep(z)=\int
dx\,\qprep(x,z)$ can be challenging.  For the former problem, we can use a
stochastic approximation, by assuming we have access to a (suitably large)
empirical distribution $\qemp(x)$.  For the latter problem, we can leverage
tractable variational bounds on mutual information
\citet{agakov2003,agakov2006variational,vib} to get  the following variational
lower and upper bounds:
\begin{equation}
	H - D \leq \Irep(X; Z) \leq R
	\label{eqn:bounds}
\end{equation}
\begin{align}
	H  &\equiv -\int dx\, \qp(x) \log \qp(x) \label{eqn:H} \\
	D  &\equiv -\int dx\, \qp(x) \int dz\, \qe(z|x) \log \qd(x|z) \label{eqn:D} \\
	R  &\equiv \int dx\, \qp(x) \int dz\, \qe(z|x) \log \frac{\qe(z|x)}{\qm(z)} \label{eqn:R}
\end{align}
where $\qd(x|z)$ (the ``\emph{decoder}'') is a variational approximation to
$\qprep(x|z)$, and $\qm(z)$ (the ``\emph{marginal}'') is a variational
approximation to $\qprep(z)$.  A detailed derivation of these bounds is
included in \Cref{sec:lowerbound,sec:upperbound}.

$H$ is the \emph{data entropy} which measures the complexity of our dataset,
and can be treated as a constant outside our control.  $D$ is the
\emph{distortion} as measured through our encoder, decoder channel, and is
equal to the reconstruction negative log likelihood.  $R$ is the \emph{rate},
and depends only on the encoder and variational marginal; it is the average
relative KL divergence between our encoding distribution and our learned
marginal approximation.  (It has this name because it measures the excess
number of bits required to encode samples from the encoder using an optimal
code designed for $\qm(z)$.)
For discrete data\footnote{If the input space is continuous, we can consider an arbitrarily fine
  discretization of the input.},
all probabilities in $X$ are bounded above by one and both the data entropy and
distortion are non-negative ($H \geq 0, D \geq 0$).  The rate is also
non-negative ($R \geq 0$), because it is an average KL divergence, for either
continuous or discrete $Z$.

%% file: phase.tex
\paragraph{Phase Diagram}

\begin{figure}[tb]
    {\includegraphics[width=0.4\textwidth]{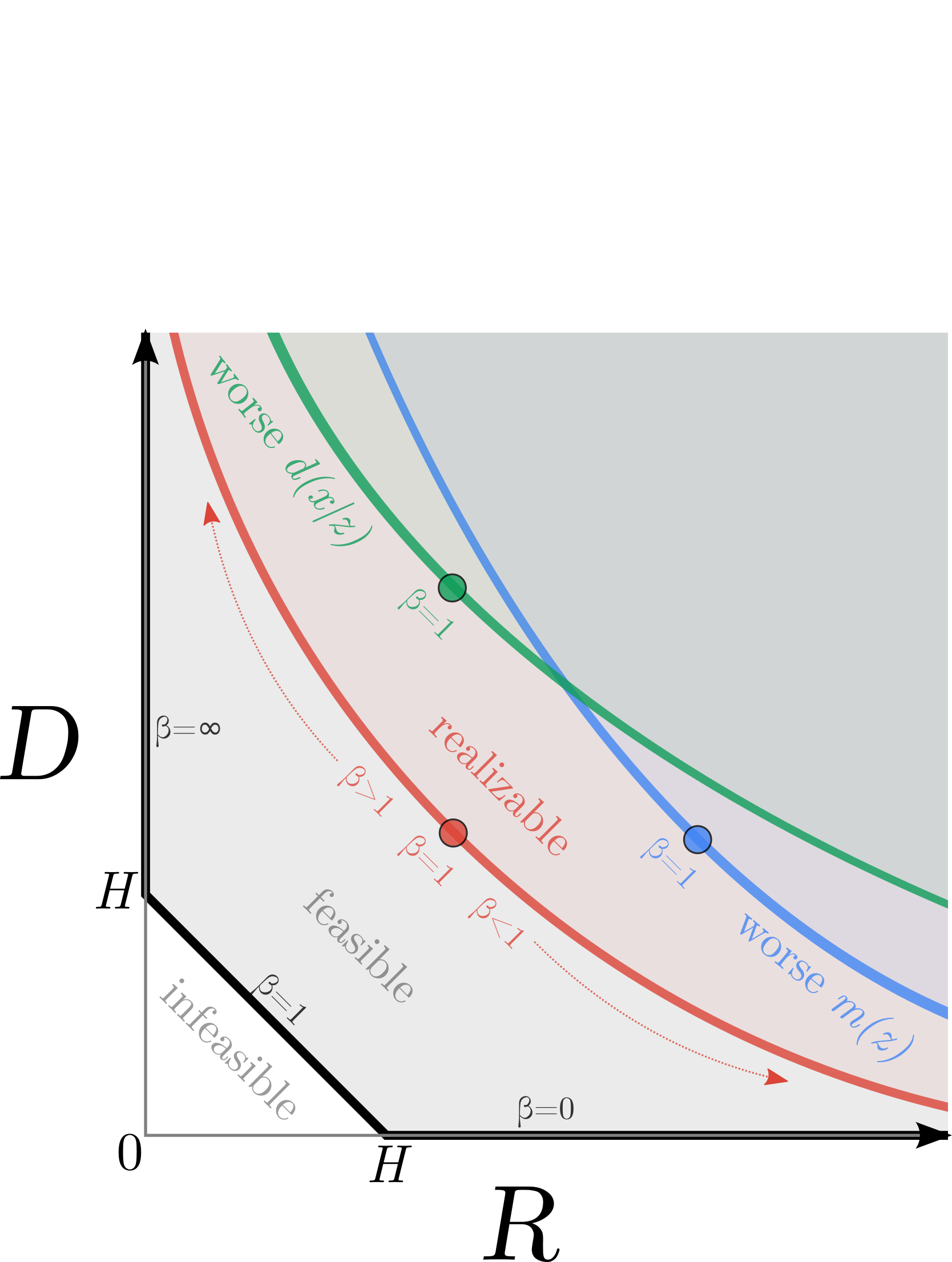}}
	\centering
  {\caption{
          %\tiny
          Schematic representation of the phase diagram in the
	$RD$-plane.  The \emph{distortion} ($D$) axis measures the
	reconstruction error of the samples in the training set.  The
	\emph{rate} ($R$) axis measures the relative KL divergence between the
	encoder and our own marginal approximation.  The thick black lines
	denote the feasible boundary in the infinite model capacity limit.
        }
      \label{fig:phasediagram}}
\end{figure}

The positivity constraints and the sandwiching bounds (\Cref{eqn:bounds})
separate the $RD$-plane into feasible and infeasible regions, visualized in
Figure~\ref{fig:phasediagram}.
The boundary between these regions is a convex curve (thick black line).
We now explain qualitatively what the different areas of
  this diagram correspond to. For simplicity,
  we will consider the infinite model family limit,
where we have complete freedom in specifying $\qe(z|x), \qd(x|z)$ and $\qm(z)$ but
consider the data distribution $\qp(x)$ fixed.

The bottom horizontal line corresponds to the zero distortion setting,
which implies that we can perfectly
encode and decode our data; we call this
the {\em auto-encoding limit}.
The lowest possible rate is given by $H$, the entropy of the data.
This corresponds to the point $(R=H,D=0)$.
(In this case, our lower bound is tight, and hence
$\qd(x|z) = \qprep(x|z)$.)
We can obtain higher rates at zero distortion, or at any other fixed distortion
by making the marginal approximation $\qm(z)$ a weaker approximation to
$\qprep(z)$, and hence simply increasing the cost of encoding our latent
variables, since only the rate and not the distortion depends on $\qm(z)$.

The left  vertical line corresponds to the zero rate setting.  Since $R = 0
\implies \qe(z|x) = \qm(z)$, we see that our encoding distribution $\qe(z|x)$
must itself be independent of $x$.  Thus the latent representation is not
encoding any information about the input and we have failed to create a useful
learned representation.  However, by using a suitably powerful decoder,
$\qd(x|z)$, that is able to capture correlations between the components of $x$
we can still reduce the distortion to the
lower bound of $H$, thus achieving the point $(R=0,D=H)$; we call this the {\em
auto-decoding limit}.
(Note that since $R$ is  an upper bound on the non-negative
mutual information, in the limit that $R = 0$, the bound must be tight, which
guarantees that $\qm(z)=\qprep(z)$.)
We can achieve solutions further up on the
$D$-axis, while keeping the rate fixed, simply by making the decoder worse,
and hence our reconstructions worse,
since only the distortion and not the rate depends on $\qd(x|z)$.

Finally, we  discuss solutions along the diagonal line. Such points satisfy
$D=H-R$, and hence both of our bounds are tight, so $\qm(z)=\qprep(z)$ and
$\qd(x|z)=\qprep(x|z)$.  (Proofs of these claims are given in
Sections~\ref{sec:optimalmarginal} and~\ref{sec:optimaldecoder} respectively.)

So far, we have considered the infinite model family limit.  If we have only
finite parametric families for each of $\qd(x|z), \qm(z), \qe(z|x)$, we expect
in general that our bounds will not be tight.  Any failure of the approximate
marginal $\qm(z)$ to model the true marginal $\qprep(z)$, or the decoder
$\qd(x|z)$ to model the true likelihood $\qprep(x|z)$, will lead to a gap with
respect to the optimal black surface.
However, our inequalities must still hold, which suggests that there will
still be a one dimensional optimal
frontier, $D(R)$, or $R(D)$ where optimality is defined to be the tightest
achievable sandwiched bound within the parametric family.
We will use the term {\em $RD$ curve} to refer to this optimal surface in the rate-distortion ($RD$)
plane.

Furthermore, by the same arguments as above, this surface should be monotonic
in both $R$ and $D$, since for any solution, with only very mild assumptions on
the form of the parametric families, we should always be able to make $\qm(z)$
less accurate in order to increase the rate at fixed distortion (see shift from
red curve to blue curve in \cref{fig:phasediagram}), or make the decoder
$\qd(x|z)$ less accurate to increase the distortion at fixed rate (see shift
from red curve to green curve in \cref{fig:phasediagram}).  Since the data
entropy $H$ is outside our control, this surface can be found by means of
constrained optimization, either minimizing the distortion at some fixed rate
(see \cref{sec:toymodel}), or minimizing the rate at some fixed distortion.

%% file: betavae.tex
\paragraph{Connection to $\beta$-VAE}

Alternatively,
instead of considering the rate as fixed, and tracing out the
optimal distortion as a function of the rate $D(R)$,
 we can perform a Legendre transformation
%(since we have established that the surface is convex),
and can find the optimal rate and distortion for a fixed
$\beta = \frac{\partial D}{\partial R}$, by minimizing
$\min_{\qe(z|x),\qm(z),\qd(x|z)} D + \beta R$.
Writing this objective out in full, we get
\begin{equation}
	\begin{split}
	\min_{\qe(z|x),\qm(z),\qd(x|z)} \int dx\, \qp(x) \int dz\, \qe(z|x) \\ \left[- \log \qd(x|z)
	+ \beta \log \frac{\qe(z|x)}{\qm(z)} \right].
	\label{eqn:betavae}
	\end{split}
\end{equation}
If we set $\beta=1$, (and identify $e(z|x) \to q(z|x), d(x|z) \to p(x|z), m(z) \to p(z)$)
this matches the ELBO objective
used when training a VAE~\citep{vae}, with the distortion term matching the
reconstruction loss, and the rate term matching the ``KL term'' ($\text{ELBO} = -(D + R)$).
Note, however, that this objective does not distinguish between any of
the points along the diagonal of the optimal $RD$ curve,
all of which have $\beta=1$ and the same ELBO\@.
Thus the ELBO objective alone (and the marginal likelihood) cannot distinguish
between models that make no use of the latent variable (autodecoders)
versus models that make
large use of the latent variable and learn useful representations for
reconstruction (autoencoders), in the infinite model family,
as noted in~\citet{huszar,mutualae}.

In the finite model family case, ELBO targets a single point along the rate
distortion curve, the point with slope 1. Exactly where this slope 1 point lies
is a sensitive function of the model architecture and the relative powers of
the encoder, decoder and marginal.

If we allow a general $\beta \geq 0$, we get the $\beta$-VAE objective used in
\citep{betavae,vib}.  This allows us to smoothly interpolate between
auto-encoding behavior ($\beta \ll 1$), where the distortion is low but the
rate is high, to auto-decoding behavior ($\beta \gg 1$), where the distortion
is high but the rate is low, all without having to change the model
architecture. Notice however that if our model family was rich enough to have a
region of its $RD$-curve with some fixed slope (e.g.\ in the extreme case, the
$\beta=1$ line in the infinite model family limit), the $\beta$-VAE objective
cannot uniquely target any of those equivalently sloped points.  In these
cases, fully exploring the frontier would require a different constraint.

%% file: related.tex
\section{Related Work}
\label{sec:related}

\paragraph{Improving VAE representations.}
Many recent papers have introduced mechanisms for alleviating the problem of
unused latent variables in VAEs.
\citet{bowman2015generating} proposed annealing the weight of the KL term of
the ELBO from 0 to 1 over the course of training but did not consider ending
weights that differed from 1. 
\citet{betavae} proposed the
$\beta$-VAE for unsupervised learning, which is a generalization
of the original VAE in which the KL term is scaled by $\beta$,
similar to this paper. However, their focus was on disentangling
and did not discuss rate-distortion tradeoffs across model families.
% beta-VAE, VLAE, infovae, infogan
Recent work has used the $\beta$-VAE objective to tradeoff reconstruction
quality for sampling accuracy \citep{ha2018a}. \citet{vlae} present a bits-back
interpretation \citep{hinton1993keeping}. %TODO: discuss/contrast?
Modifying the variational families \citep{iaf},
priors \citep{maf,vampprior}, and decoder structure \citep{vlae}
have also been proposed as a mechanism for learning better representations.

\paragraph{Information theory and representation learning.}
The information bottleneck
framework leverages information theory to learn robust
representations~\citep{Tishby99, generalization, deeplearningib, vib,
  infodropout, emergence}. It allows a
model to smoothly trade off the minimality of the learned
representation ($Z$) from data ($X$) by minimizing their mutual
information, $I(X;Z)$, against the informativeness of the
representation for the task at hand ($Y$) by maximizing their mutual
information, $I(Z;Y)$. \citet{deeplearningib}
plot an RD curve similar to the one in this paper, but they only
consider the supervised setting.
%and they do not consider the information content
%that is implicit in powerful stochastic decoders.

Maximizing mutual information to power unsupervised representational learning has
a long history. \citet{blindsource} uses an information maximization
objective to derive the ICA algorithm for blind source
separation. \citet{unsupib} learns clusters with the Blahut-Arimoto algorithm.
\citet{agakov2003} was the first
to introduce tractable variational bounds on mutual information,
and made close analogies and comparisons to maximum
likelihood learning and variational autoencoders. Recently, information theory
has been useful for reinterpreting the ELBO \citep{surgery},  and understanding
the class of tractable objectives for training generative models
\citep{infoautoencoding}.

Recent work has also presented information maximization as a solution to the 
problem of VAEs ignoring the latent code. \citet{infovae} modifies the ELBO by replacing the
rate term with a divergence from the aggregated posterior to the prior and
proves that solutions to this objective maximize the representational mutual
information.
However, their objective requires leveraging techniques from implicit variational inference as the
aggregated posterior is intractable to evaluate. \citet{infogan} also presents
an approach for maximizing information but requires the use of adversarial
learning to match marginals in the input space.
Concurrent work from \citet{mutualae} present a similar framework for
maximizing information in a VAE through a variational lower bound on the
generative mutual information. Evaluating this bound requires sampling the
generative model (which is slow for autoregressive models) and computing
gradients through model samples (which is challening for discrete input
spaces). In Section~\ref{sec:toymodel}, we present a similar approach that uses
a tractable bound on information that can be applied to discrete input spaces
without sampling from the model.

\paragraph{Generative models and compression.}
Rate-distortion theory has been used in compression to tradeoff
the size of compressed data with the fidelity of the reconstruction.
Recent approaches to compression have leveraged deep latent-variable generative
models for images, and explored tradeoffs in the RD plane
\citep{conceptualcompression,endtoendcompression, johnston2017improved}.
However, this work focuses on a restricted set of architectures with simple
posteriors and decoders and does not study the impact that architecture choices
have on the marginal likelihood and structure of the representation.

%% file: experiments.tex
\section{Experiments}
\label{sec:experiments}

	\InputIfFileExists{toymodel}{}{\typeout{No file toymodel.}}%

%
	\InputIfFileExists{staticmnist}{}{\typeout{No file staticmnist.}}%

	\InputIfFileExists{omniglot}{}{\typeout{No file omniglot.}}%

%% file: toymodel.tex
\paragraph{Toy Model}
\label{sec:toymodel}

In this section, we empirically show a case where the usual ELBO objective can
learn a model which perfectly captures the true data distribution, $\qp(x)$,
but which fails to learn a useful latent representation.  However, by training
the {\em same model} such that we minimize the distortion, subject to achieving
a desired target rate $R^*$, we can recover a latent representation that
closely matches the true generative process (up to a reparameterization), while
also perfectly capturing the true data distribution.
In particular,
we solve the following optimization problem:
$\min_{\qe(z|x),\qm(z),\qd(x|z)} D + |\sigma - R|$ where $\sigma$ is the target rate.
(Note that, since we use very flexible nonparametric models, we can achieve
$\qprep(x)=\qp(x)$ while ignoring $z$, so using the $\beta$-VAE approach would not suffice.)

We create a simple data generating process that consists of a true latent
variable $Z^*=\{z_0,z_1\} \sim \textrm{Ber}(0.7)$ with added Gaussian noise and
discretization.  The magnitude of the noise was chosen so that the true
generative model had $\MI(x;z^*) = 0.5$ nats of mutual information between the
observations and the latent.  We additionally choose a model family with
sufficient power to perfectly autoencode or autodecode.  See
Appendix~\ref{sec:toymodel-extra} for more detail on the data generation and
model.

\begin{figure*}[tbp]
	\centering
	\subfloat[Optimal (hand-constructed)]{
		\includegraphics[width=0.337\textwidth]{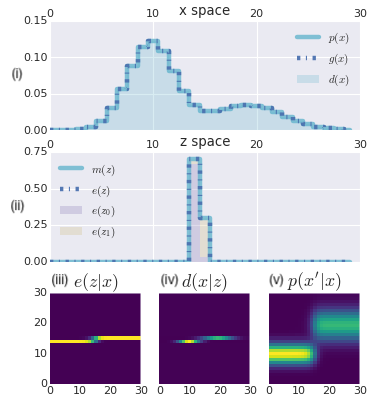}
		\label{fig:toymodelinv}
	}
	\subfloat[VAE]{
		\includegraphics[width=0.32\textwidth]{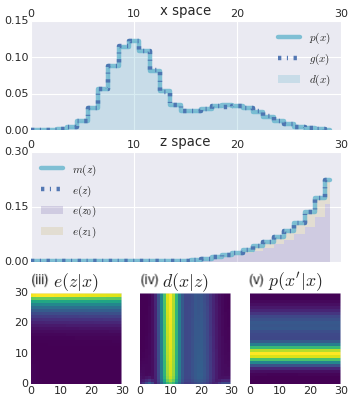}
		\label{fig:toymodelelbo}
	}
	\subfloat[Target Rate]{
		\includegraphics[width=0.32\textwidth]{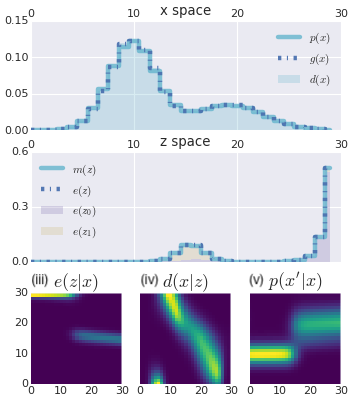}
		\label{fig:toymodelrate}
	}

	\caption{
        \small
        Toy Model illustrating the difference between fitting
          a model by maximizing ELBO (b)
	  vs minimizing distortion for a fixed rate (c).
	  {\bf Top (i):} Three distributions in data space: the true data
          distribution, $\qp(x)$, the model's generative distribution, $g(x) = \sum_z
          \qm(z) \qd(x|z)$, and the empirical data reconstruction distribution, $d(x) =
          \sum_{x'} \sum_z \qemp(x') \qe(z|x') \qd(x|z)$.
	  {\bf Middle (ii):} Four distributions in latent space: the learned (or
          computed) marginal $\qm(z)$, the empirical induced marginal $\qe(z) = \sum_x
          \qemp(x) \qe(z|x)$, the empirical distribution over $z$ values for data vectors
          in the set $\calX_0 = \{ x_n: z_n = 0\}$, which we denote by $e(z_0)$ in
          purple, and the empirical distribution over $z$ values for data vectors in the
          set $\calX_1 = \{ x_n: z_n = 1\}$, which we denote by $e(z_1)$ in yellow.
	  {\bf Bottom:} Three $K \times K$ distributions: (iii) $\qe(z|x)$, (iv) $\qd(x|z)$ and
	  (v) $p(x'|x)=\sum_z \qe(z|x) \qd(x'|z)$.
	}
        \label{fig:toymodel}
\end{figure*}

Figure~\ref{fig:toymodel} shows various distributions computed using three
models.  For the left column (\ref{fig:toymodelinv}), we use a hand-engineered encoder $\qe(z|x)$,
decoder $\qd(x|z)$, and marginal $\qm(z)$ constructed with knowledge of the
true data generating mechanism to illustrate an optimal model.
For the middle (\ref{fig:toymodelelbo}) and right (\ref{fig:toymodelrate}) columns, we learn $\qe(z|x)$, $\qd(x|z)$, and $\qm(z)$
using effectively infinite data sampled from $\qp(x)$ directly.
The middle column (\ref{fig:toymodelelbo}) is trained with ELBO\@. The right column (\ref{fig:toymodelrate}) is trained by targeting
$R=0.5$ while minimizing $D$.\footnote{Note that the target value $R = \MI(x;z^*)=0.5$ is computed with knowledge of the true data
  generating distribution. However, this is the only information that is
  ``leaked'' to our method, and in general it is not hard to guess reasonable
  targets for $R$ for a given task and dataset.}
In both cases, we see that $\qp(x) \approx g(x) \approx \qd(x)$ for both trained models (2bi, 2ci),
indicating that optimization found the global optimum of the respective objectives.
However, the VAE fails to learn a useful representation, only
yielding a rate of $R=0.0002$ nats,\footnote{This is an example of VAEs ignoring the latent space. As decoder power increases,
  even $\beta=1$ is sufficient to cause the model to collapse to the autodecoding limit.}
while the Target Rate model achieves $R=0.4999$ nats. Additionally, it nearly perfectly reproduces the
true generative process, as can be seen by comparing the yellow and purple regions in the
z-space plots (2aii, 2cii) -- both the optimal model and the Target Rate model have two clusters,
one with about 70\% of the probability mass, corresponding to class 0 (purple shaded region),
and the other with about 30\% of the mass (yellow shaded region) corresponding to class 1.
In contrast, the z-space of the VAE (2bii) completely mixes the yellow and purple regions, only
learning a single cluster. Note that we reproduced essentially identical results with dozens of
different random initializations for both the VAE and the penalty VAE model -- these results are
not cherry-picked.

%% file: staticmnist.tex
%\paragraph{Static Binary MNIST and OMNIGLOT}
\paragraph{MNIST: $RD$ curve}%
In this section, we show how comparing models in terms of rate and distortion separately is
more useful than simply observing marginal log likelihoods, and allows a detailed ablative
comparison of individual architectural modifications.
We use the static binary MNIST dataset from
\citet{binarystaticmnist}\footnote{\url{https://github.com/yburda/iwae/tree/master/datasets/BinaryMNIST}}.

We examine several VAE model architectures that have been proposed in the literature.
In particular,
we  consider simple and complex variants for the encoder and decoder, and
three different types of marginal.  The simple encoder is a CNN  with a fully factored 64 dimensional
Gaussian for $\qe(z|x)$; the more complex encoder is similar, but followed by 4
steps of mean-only Gaussian inverse autoregressive flow~\citep{iaf}, with each
step implemented as a 3 hidden layer MADE~\citep{made} with 640 units in each
hidden layer.  The simple decoder is a multilayer deconvolutional network; the
more powerful decoder is a PixelCNN++~\citep{pxpp} model.  The simple marginal
is a fixed isotropic Gaussian, as is commonly used with VAEs; the more complicated version has a 4 step 3
layer MADE~\citep{made} mean-only Gaussian autoregressive flow~\citep{maf}.
We also consider the setting in
which the marginal uses the VampPrior from \citep{vampprior}.
We will denote the particular model combination by the tuple $(+/-, +/-, +/-/v)$,
depending on whether we use a  simple $(-)$ or complex $(+)$ (or $(v)$ VampPrior)
version for the (encoder, decoder, marginal) respectively.
In total we consider $2 \times 2 \times 3 = 12$ models.  We train them all to minimize the
$\beta$-VAE objective in Equation~\ref{eqn:betavae}.  Full details can be
found in Appendix~\ref{sec:mnistdetails}.
Runs were performed at various values of $\beta$ ranging from 0.1 to 10.0, both
with and without KL annealing~\citep{bowman2015generating}.

\begin{figure*}[tb]
	\centering
  \subfloat[Distortion vs Rate]{\includegraphics[width=0.47\textwidth]{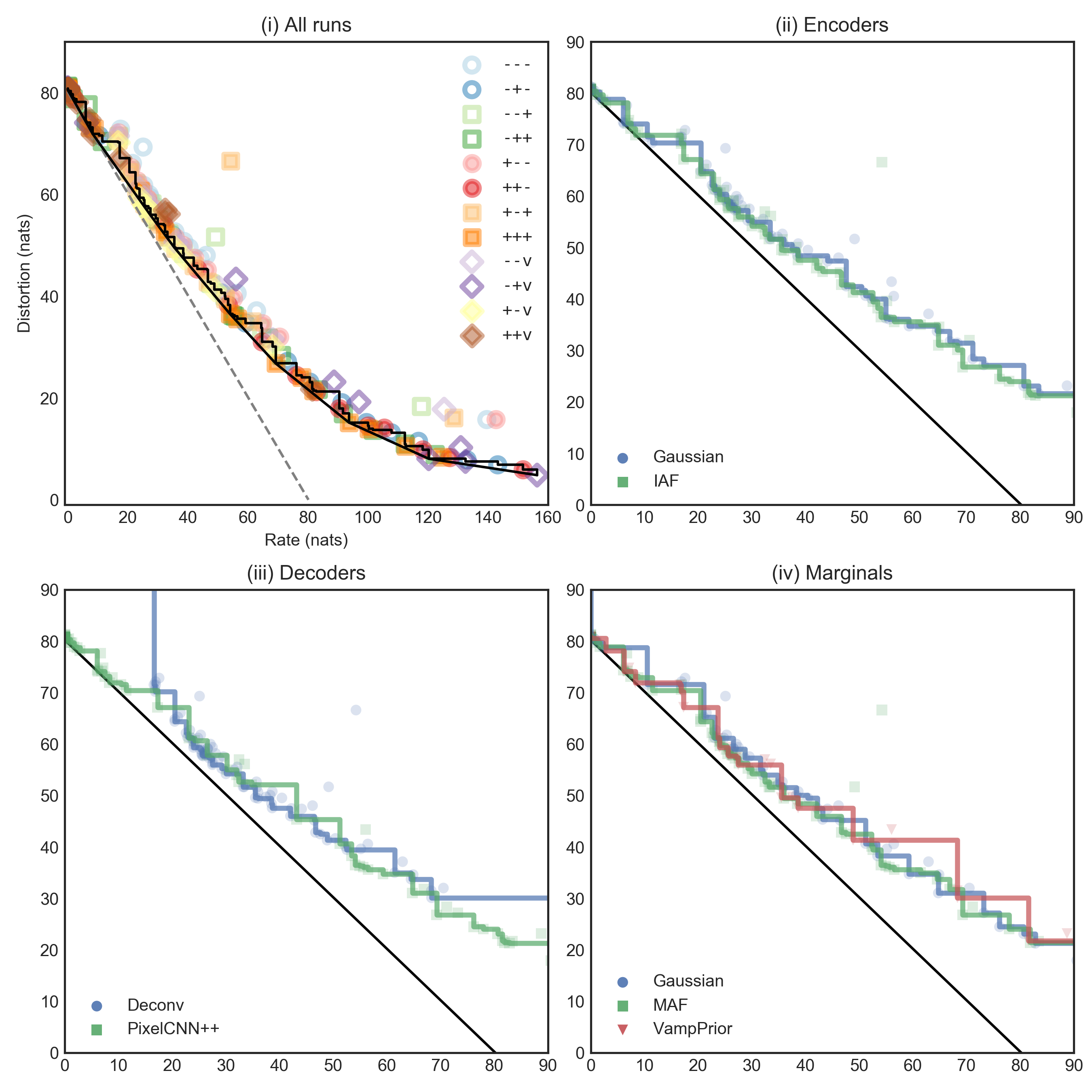}
	\label{fig:staticmnistrd}}
	\subfloat[ELBO ($R+D$) vs Rate]{\includegraphics[width=0.47\textwidth]{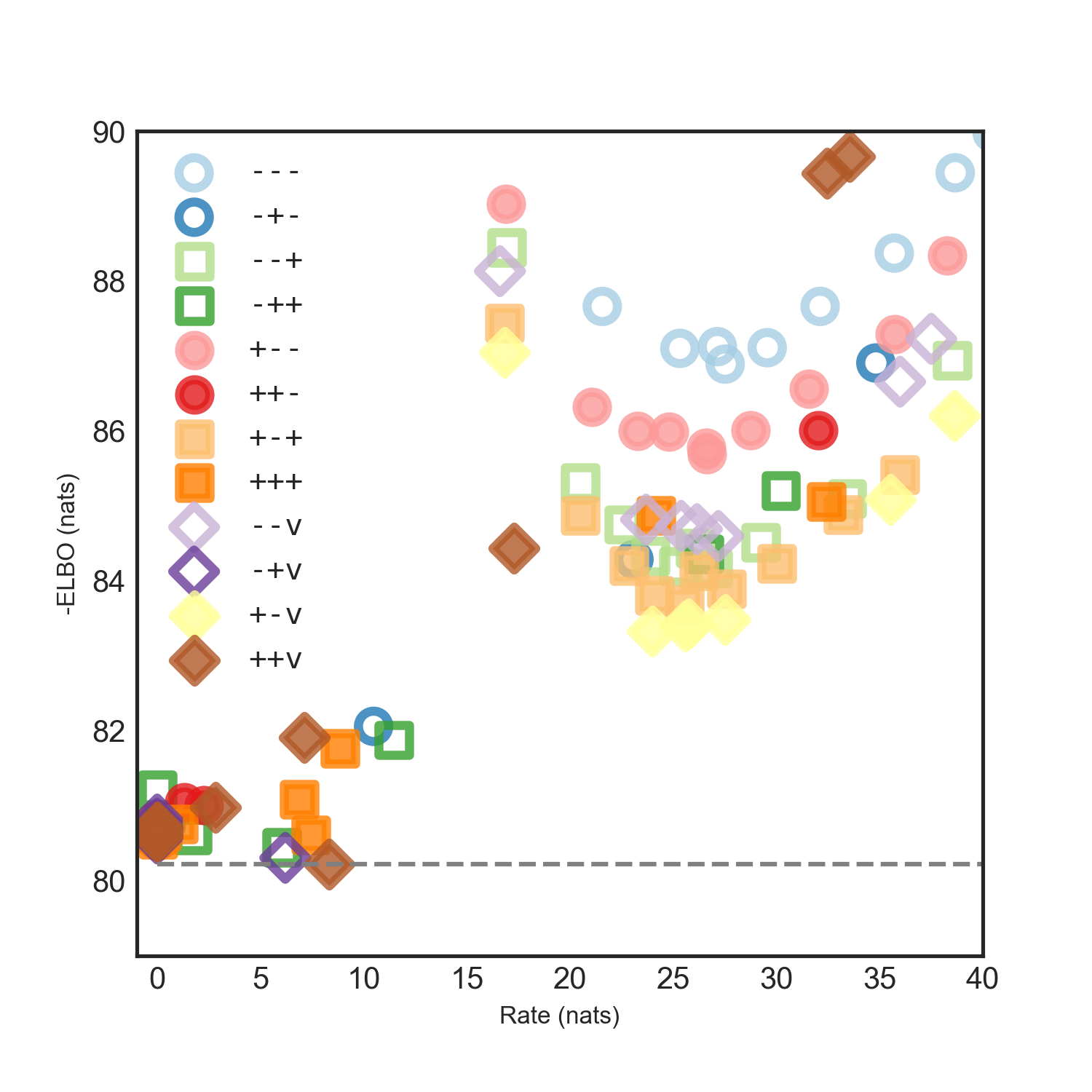}
	\label{fig:staticmnistelbor}}
  \caption{
  Rate-distortion curves on MNIST.
  \small
 (a) We plot the best $(R,D)$ value obtained by various models,
 denoted by the tuple $(e,d,m)$,
 where
 $e \in \{-,+\}$ is the simple Gaussian or complex IAF encoder,
  $d \in \{-,+\}$ is the simple deconv or complex pixelCNN++ decoder,
 and $m \in \{-,+,v)$ is the simple Gaussian, complex MAF or even more
 complex Vamp marginal.
 The top left shows all architectures individually.  The next three panels show the computed
frontier as we sweep $\beta$ for a given pair (or triple) of model
  types.
(b) The same data, but on the skew axes of
  -ELBO = $R+D$ versus $R$.
Shape encodes the
marginal, lightness of color denotes the decoder, and fill denotes the
encoder.
  }
  \label{fig:mnistRD}
\end{figure*}

%\paragraph{RD curve.}%
\Cref{fig:staticmnistrd}(i) shows the converged $RD$ location for a total of
209 distinct runs across our 12 architectures, with different initializations
and $\beta$s on the  MNIST dataset.
The best ELBO we achieved was $\hat{H} = 80.2$ nats,
at $R=0$.
This  sets an upper bound on the true data entropy
$H$ for the static MNIST dataset.
The dashed line connects $(R=0,D=\hat{H})$ to $(R=\hat{H},D=0)$,
This implies that any $RD$ value above the
dashed line is in principle achievable in a powerful enough model.  The
stepwise black curves show the monotonic Pareto frontier of achieved $RD$
points across all model families.  The grey solid line shows the
corresponding convex hull, which we approach closely across all rates.
% Strong
% decoder model families dominate at the lowest and highest rates.  Weak decoder
% models dominate at intermediate rates. Strong marginal models dominate strong
% encoder models at most rates.  Across our model families we appear to be
% pushing up against an approximately smooth $RD$ curve.
The 12 model families
we considered here, arguably a representation of the classes of models
considered in the VAE literature, in general perform much worse in the
auto-encoding limit (bottom right corner) of the $RD$ plane. This is likely due
to a lack of power in our current marginal approximations, and suggests
more experiments with powerful autoregressive marginals, as in \citet{vqvae}.

\Cref{fig:staticmnistrd}(iii) shows the same data, but this time focusing
on the conservative Pareto frontier across all architectures with either a
simple deconvolutional decoder (blue) or a complex autoregressive decoder
(green). Notice the systematic failure of simple decoder models at the lowest
rates.  Besides that discrepancy, the frontiers largely track one another at
rates above ~22 nats.
This is perhaps unsurprising considering we
trained on the binary MNIST dataset, for which the measured pixel level
sampling entropy on the test set is approximately 22 nats.
When  we plot the same data where we vary the encoder (ii) or marginal (iv)
from simple to complex,  we do not see any systematic trends.
%Making the same comparison looking only at (ii)
%encoders, or (iv) marginals, it is hard to argue that it is any one of
%these architectural improvements alone that creates the best models.
\Cref{fig:staticmnistelbor} shows the same raw data, but we plot
-ELBO=$R+D$ versus $R$.  Here some of the differences between individual model
families' performances are more easily resolved.

\begin{figure*}[tb]
	\centering
	\subfloat[Reconstructions from \texttt{-+v} with $\beta=0.1-1.1$.]
        {\includegraphics[height=0.28\textwidth]{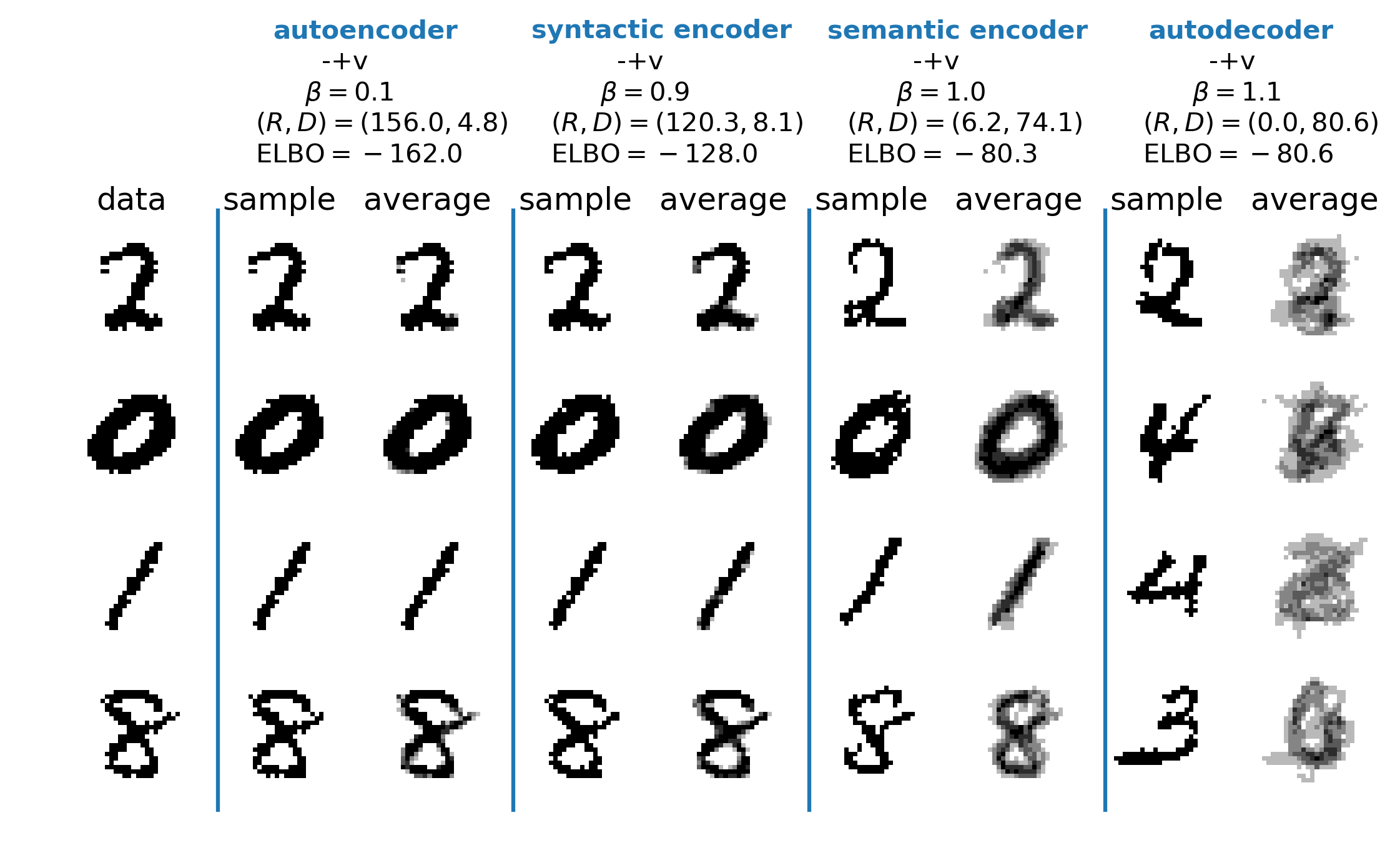}\quad
	\label{fig:mnistrecon}}
	\subfloat[Generations from \texttt{-+v} with $\beta=0.1-1.1$]
        {\includegraphics[height=0.28\textwidth]{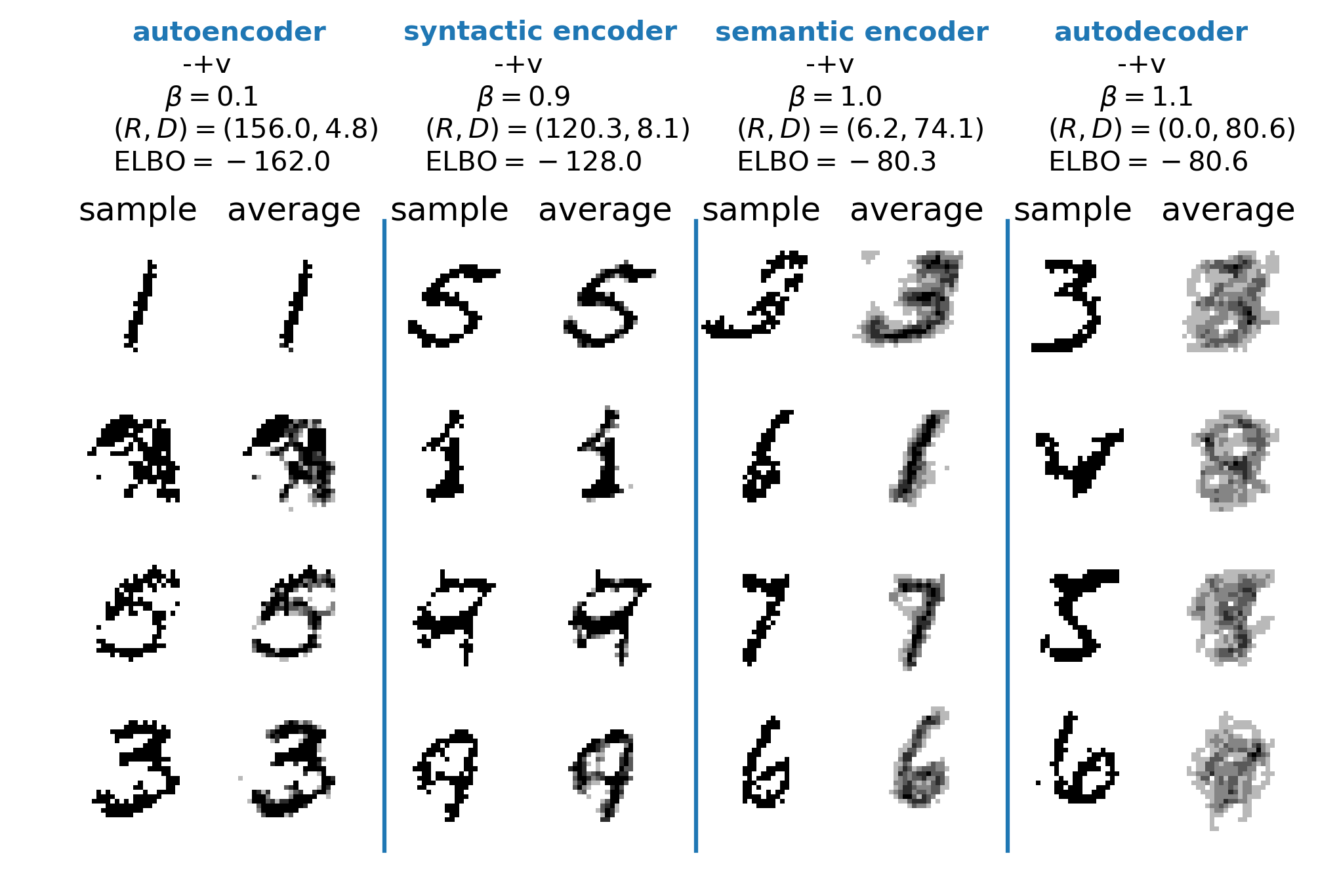}
	\label{fig:mnistgen}}

   \subfloat[Reconstructions from 4 VAE models with $\beta=1$.]
   {\includegraphics[height=0.28\textwidth]{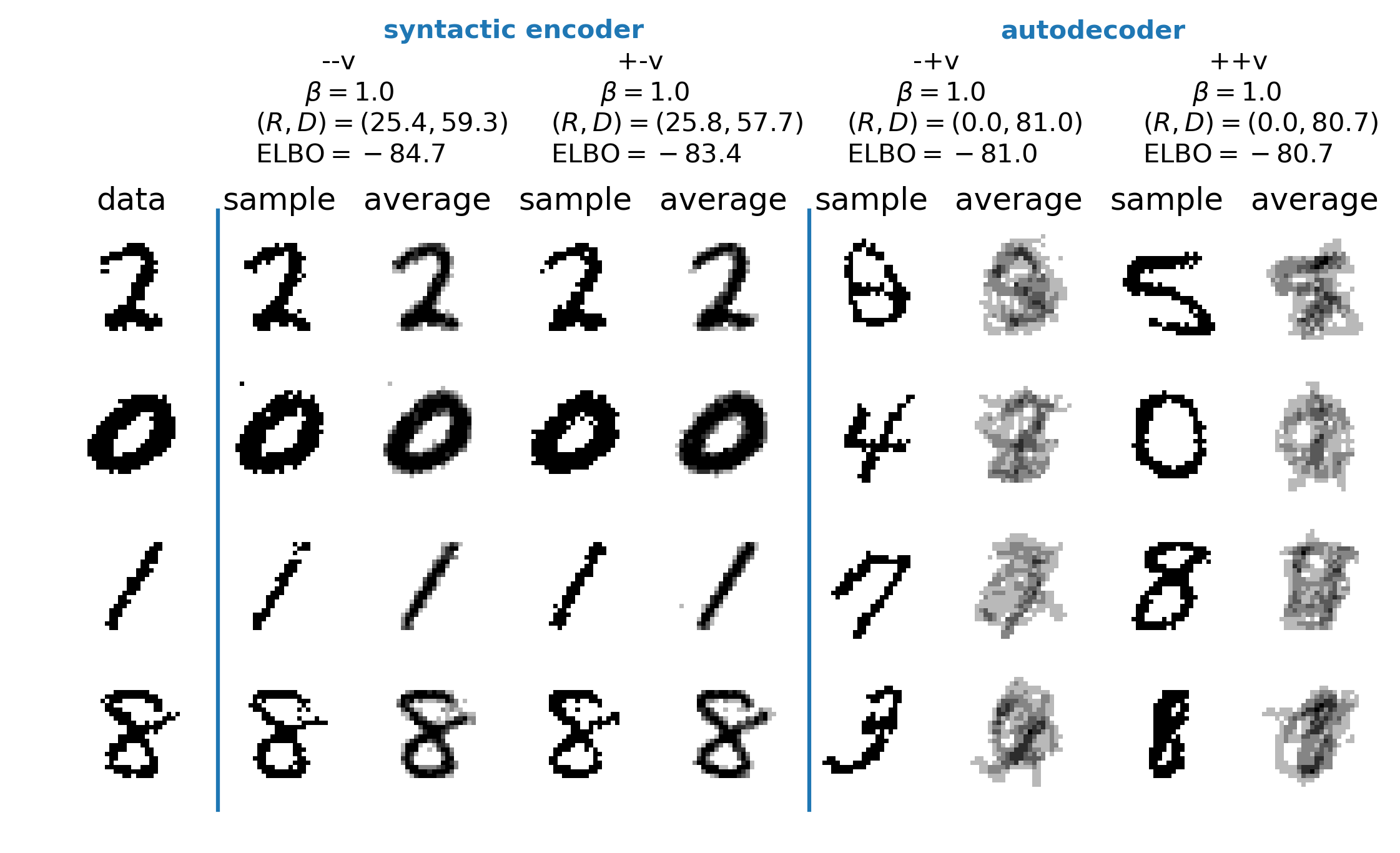}
 	\label{fig:mnistb1recon}}
   \subfloat[Reconstructions from models with the same ELBO.]
   {\includegraphics[height=0.28\textwidth]{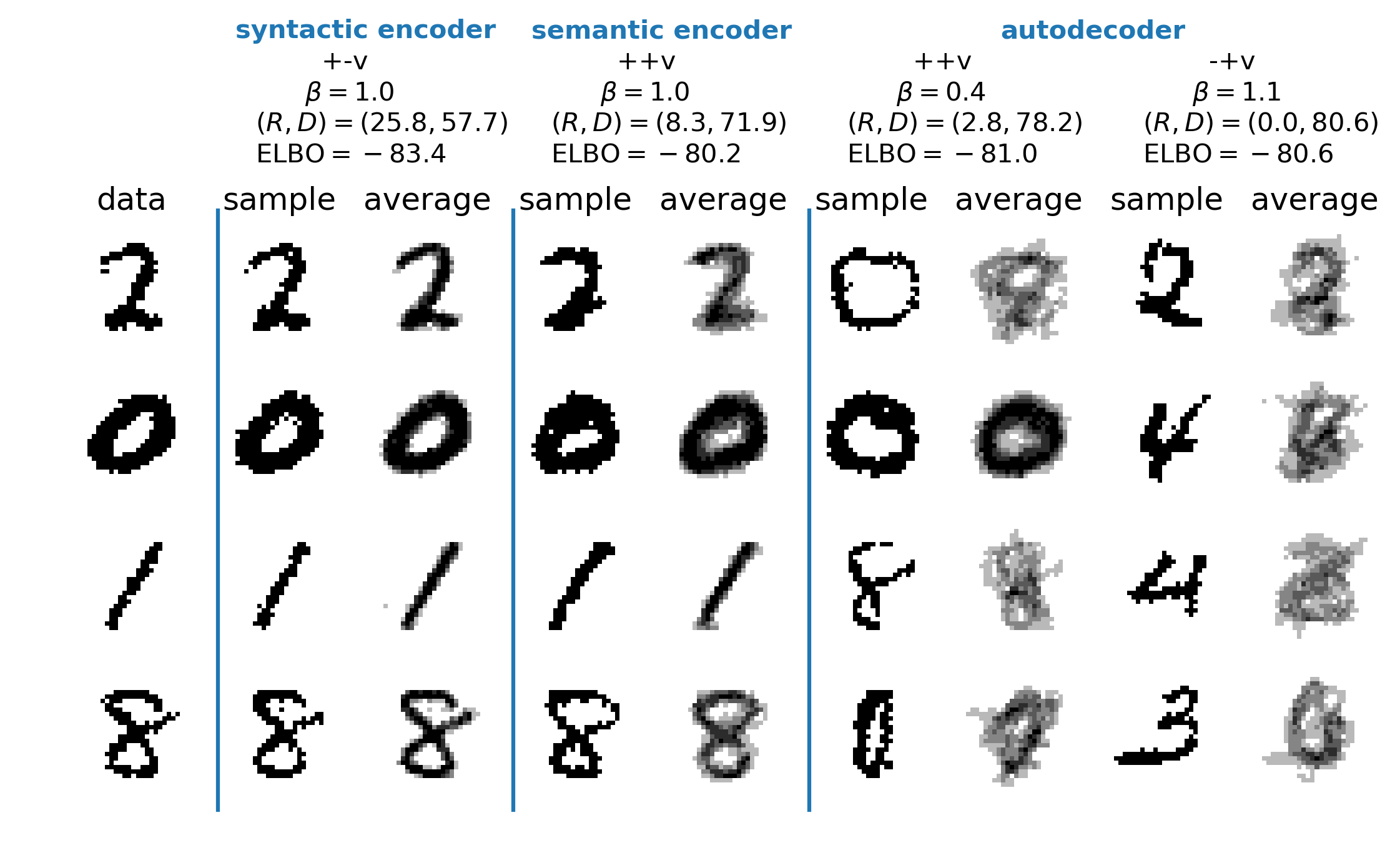}
 	\label{fig:mnistfrontierrecon}}

  \caption{
  \small
	Here we show sampled reconstructions
 $z \sim e(z|x)$, $\hat{x} \sim d(x|z)$
and generations
$z \sim m(z)$, $\hat{x} \sim d(x|z)$
from various model configurations.
Each row is a different sample.
Column `data' is the input for reconstruction.
Column `sample' is a single binary image sample.
Column `average' is the mean of 5 different samples of the decoder holding the encoding $z$ fixed.
        (a-b) By adjusting $\beta$ in a fixed model architecture, we can
	smoothly interpolate between nearly perfect autoencoding on the left and nearly perfect autodecoding on the right.
  In between the two extremes are examples of syntactic encoders and semantic encoders.
	(c) By fixing $\beta=1$ we see the behavior of different architectures when trained as traditional VAEs.
        Here only 4 architectures are shown but the sharp transition from syntactic encoding on the left to autodecoding on the right is apparent.
        At $\beta=1$, only one of the 12 architectures achieved semantic encoding.
	The complete version is in \Cref{fig:vaeimages} in the Appendix.
	(d) Here we show a set of models all with similar, competative ELBOs.
         While these models all have similar ELBOs, their qualitative
  performance is very different, again smoothly interpolating between the
  perceptually good reconstructions of the syntactic decoder, the syntactic variation of the semantic encoder, and finally two clear autodecoders.
	A more complete trace can be found at \Cref{fig:frontier}.
  See text for discussion.
  }
  \label{fig:mnistSamples}
\end{figure*}

\paragraph{MNIST: Samples}
To qualitatively evaluate model performance,  \Cref{fig:mnistSamples}
 shows sampled reconstructions and generations from some of the
 runs, which we have grouped into rough categories: {\em autoencoders},
 {\em syntactic encoders}, {\em semantic encoders}, and {\em autodecoders}.
For reconstruction,
we pick an image $x$ at random, encode it using $z \sim e(z|x)$,
and then reconstruct it using $\hat{x} \sim d(x|z)$.
For generation,
we sample $z \sim m(z)$, and then decode it
using $x \sim d(x|x)$.
In both cases, we use  the same $z$ each time we sample $x$,
in order to illustrate the stochasticity implicit in the decoder.
This is particularly important to do when using powerful decoders, such
as autoregressive models.

In \Cref{fig:mnistrecon,fig:mnistgen},
we study the effect of changing $\beta$ (using KL annealing from low to high)
on the same -+v model,
corresponding to a VAE with a simple encoder,
a powerful PixelCNN++ decoder,
and a powerful VampPrior marginal.
%In \cref{fig:mnistrecon} we assess how well the models do at reconstructing
%their inputs.

\begin{compactitem}
\item When $\beta=1.10$ (right column), the model obtains $R = 0.0004$,
$D=80.6$, ELBO=-80.6 nats, which is an example of an autodecoder.  The
tiny rate indicates that the decoder ignores its latent code, and hence the
reconstructions are independent of the input $x$.  For example, when the input
is $x=8$ (bottom row), the reconstruction is $\hat{x}=3$.
However, the generated images
in \cref{fig:mnistgen}
sampled from the decoder look good.
This is an example of an  autodecoder.

\item When $\beta=0.1$ (left column),
the model obtains $R=156,
D=4.8$ , ELBO=-161 nats.  Here the model is an excellent autoencoder,
generating nearly pixel-perfect reconstructions.  However, samples from this
model's prior, as shown
in \cref{fig:mnistgen}, are very poor quality,
which is also reflected in the worse ELBO.
This is an example of an  autoencoder.

\item When $\beta = 1.0$, (third column),
we get $R=6.2, D=74.1$, ELBO=-80.3.
This model seems to retain semantically meaningful
information about the input, such as its class and width of the strokes, but
maintains syntactic variation in the individual reconstructions, so we call this a semantic encoder.
In particular, notice
that the input ``2'' is reconstructed as a similar ``2'' but with
a visible loop at the bottom (top row).
This model also has very good generated samples.
This semantic encoding arguably typifies what we want to achieve in
unsupervised learning: we have learned a highly compressed representation that
retains semantic features of the data.
We therefore call it a ``semantic encoder''.

\item  When  $\beta=0.15$ (second column),
we get  $R=120.3, D=8.1$, ELBO=-128.
This model retains both semantic and syntactic information,
where each digit's style is maintained,
and also has a good degree of compression. We call this a
``syntactic encoder''.
However, at these higher rates the failures of
our current architectures to approach their theoretical performance becomes
more apparent, as the corresponding ELBO of 128 nats is much higher than the 81
nats we obtain at low rates.  This is also evident in the visual degradation in
the generated samples (\Cref{fig:mnistgen}).
\end{compactitem}
\eat{
While it is popular to visualize both the reconstructions and generated samples
from VAEs, we suggest researchers visually compare several sampled decodings
using the same sample of the latent variable, whether it be from the encoder or
the prior, as done here in \Cref{fig:mnistSamples}.  By using a single sample
of the latent variable, but decoding it multiple times, one can visually
inspect what features of the input are captured in the observed value for the
rate.
This is particularly important to do when using powerful decoders, such
as autoregressive models.
}

\Cref{fig:mnistb1recon} shows what happens
when we vary the model for a fixed value of $\beta=1$,
as in traditional VAE training.
Here only 4
architectures are shown (the full set is available
in \Cref{fig:vaeimages} in the appendix), but
the pattern is apparent: whenever we use a powerful decoder,
the  latent code is independent of the input,
so it cannot reconstruct well.
However,
\Cref{fig:mnistrecon} shows that by using $\beta < 1$,
we can force such models to do well at reconstruction.
Finally, \Cref{fig:mnistfrontierrecon}
shows 4 different models, chosen from the Pareto frontier,
which all have almost identical ELBO scores,
but which
exhibit qualitatively different behavior.

\eat{
Models with a powerful autoregressive decoder perform well at
low rates, but for values of $\beta\ge1$ tend to collapse to pure autodecoding
models.  \Cref{fig:mnistb1recon} demonstrates this by showing the result of
using traditional VAE training as we vary the architecture. Here only 4
architectures are shown (the full set is available
in \Cref{fig:vaeimages} in the appendix), but
the pattern is apparent.
Our framework helps explain the
observed difficulties in the literature of training a useful VAE with a
powerful decoder, and the observed utility of techniques like ``free
bits''~\citep{iaf}, ``soft free bits''~\citep{vlae} and KL
annealing~\citep{bowman2015generating}.  Each of these effectively trains at a
reduced $\beta$, moving up along the $RD$ curve.  Without any additional
modifications, simply training at reduced $\beta$ is a simpler way to achieve
nonvanishing rates, without additional architectural adjustments like in the
variational lossy autoencoder~\citep{vlae}.

Analyzing model performance using the $RD$ curve gives a
much more insightful comparison of relative model performance than simply
comparing marginal data log likelihoods, or ELBO.
In particular, we managed to
achieve models with five-sample IWAE \citep{iwae}
estimates below 82 nats (a competitive rate for single layer latent variable models~\citep{vampprior})
for rates spanning from $10^{-4}$ to 30 nats.
While all of those models have competitive ELBOs or
marginal log likelihood, they differ substantially in the trade-offs they make
between rate and distortion, and those differences result in qualitatively different model behavior,
as illustrated in~\Cref{fig:mnistfrontierrecon,fig:frontier}.
}

%% file: omniglot.tex
\paragraph{Omniglot}

We repeated the experiments on the omniglot dataset,
and find qualitatively similar results.
See \cref{sec:omniExtra} for details.

%% file: concl.tex
\section{Discussion and further work}
\label{sec:concl}

We have presented a theoretical framework for understanding representation
learning using latent variable models in terms of the rate-distortion tradeoff.
This constrained optimization problem allows us to fit models by targeting a
specific point on the $RD$ curve, which we cannot do using the $\beta$-VAE
framework.

In addition to the theoretical contribution, we have conducted a large set of
experiments, which demonstrate the tradeoffs implicitly made by several
recently proposed VAE models.  We confirmed  the power of autoregressive
decoders, especially at low rates.  We also confirmed that models with
expressive decoders can ignore the  latent code, and proposed a simple solution
to this problem (namely reducing the KL penalty term to $\beta < 1$).  This fix
is much easier to implement than other solutions that have been proposed in the
literature, and comes with a clear theoretical justification.  Perhaps our most
surprising finding is that all the current approaches seem to have a hard time
achieving high rates at low distortion.  This suggests the need to develop
better marginal posterior approximations, which should in principle be able to
reach the autoencoding limit, with vanishing distortion and rates approaching
the data entropy.

Finally, we strongly encourage future work to report rate and distortion values
independently, rather than just reporting the log likelihood, which fails to
distinguish qualitatively different behavior of certain models.

%% file: mnistextra.tex
\section{More results on Static MNIST}
\label{sec:mnistExtra}

\begin{figure}[htbp]
	\centering
	\subfloat[(reconstructions)]{\includegraphics[width=1.0\linewidth]{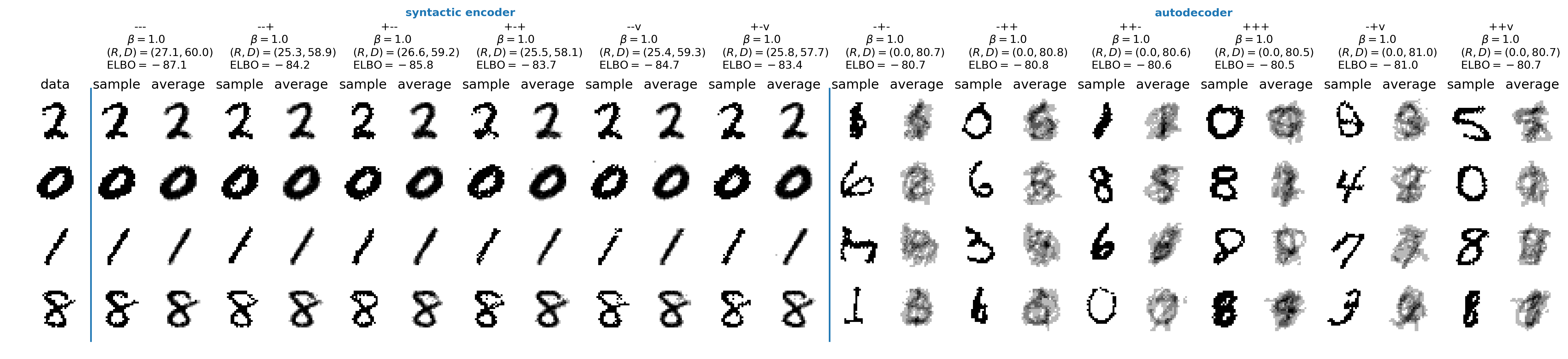}} \\
	\subfloat[(generations)]{\includegraphics[width=1.0\linewidth]{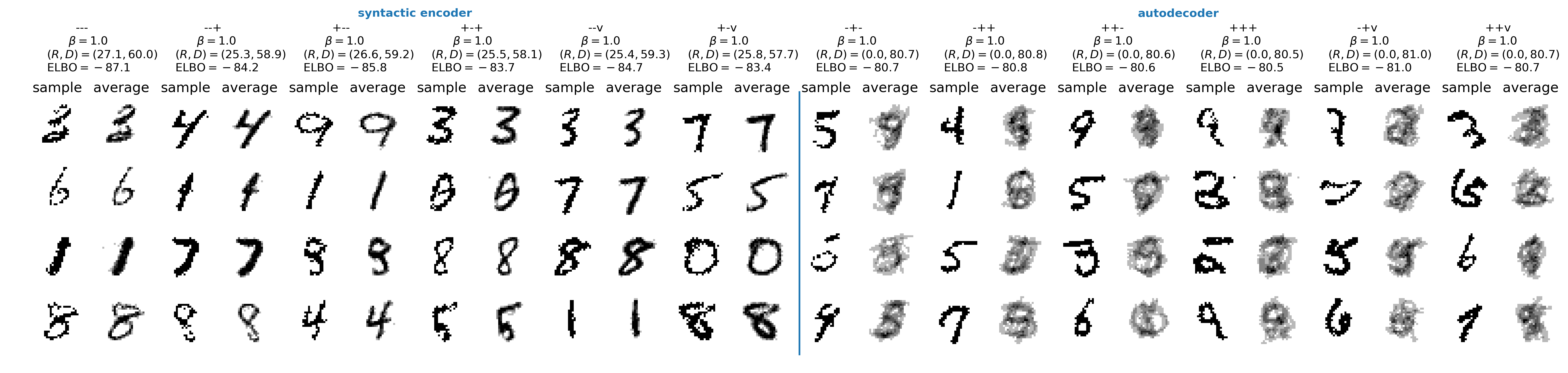}} \\

	\caption{Traditional VAE behaviors of all model families. Note the clear separation between syntactic encoders
	and autodecoders, both in terms of the rate-distortion tradeoff, and in qualitative terms of sample variance. Also
	note that none of the 12 VAEs is a semantic encoder. Semantic encoding seems difficult to achieve at $\beta=1$.
	}\label{fig:vaeimages}
\end{figure}

\begin{figure}[htbp]
	\centering
	\subfloat[(reconstructions)]{\includegraphics[width=1.0\linewidth]{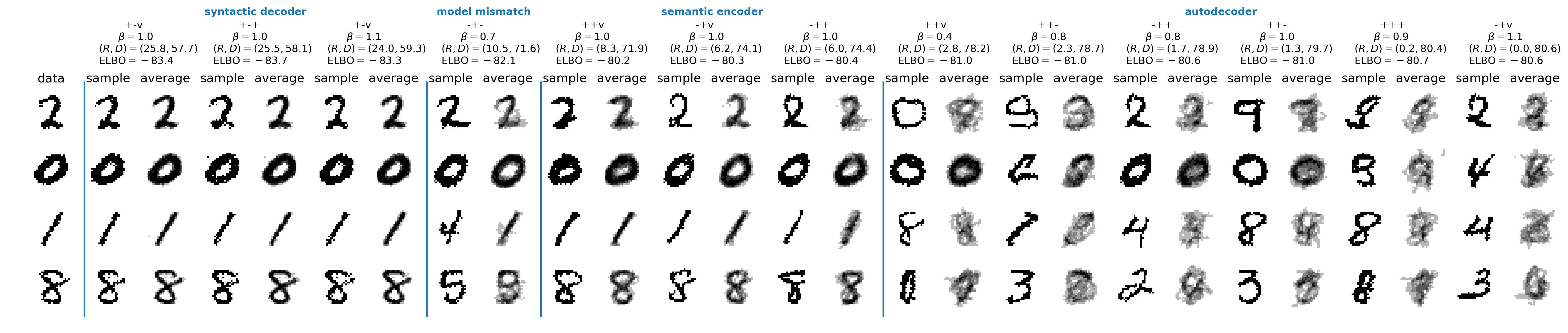}} \\
	\subfloat[(generations)]{\includegraphics[width=1.0\linewidth]{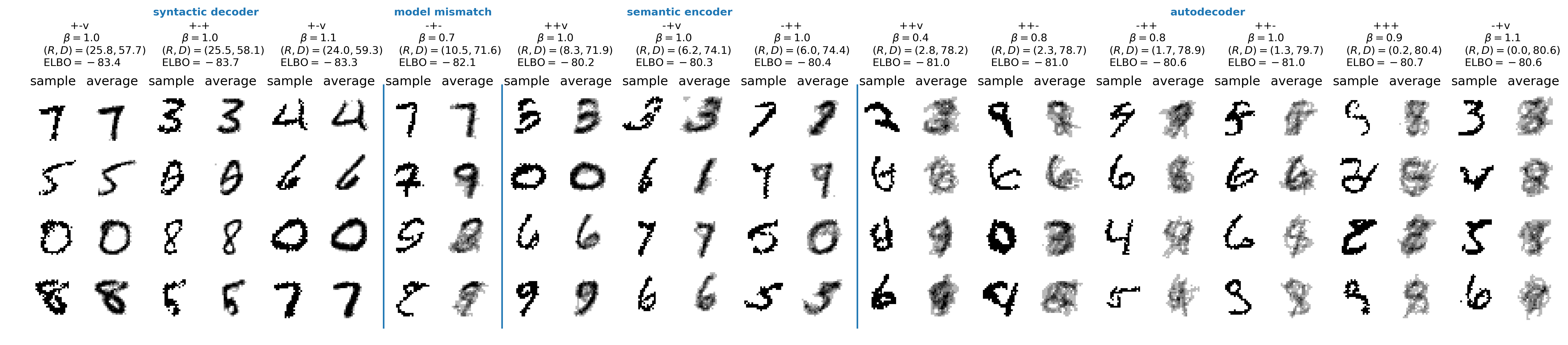}} \\
	\caption{Exploring the frontier. Here we show the reconstructions (a) and generated samples (b) from a collection of runs that all lie
	on the frontier of realizable rate distortion tradeoffs.
	}\label{fig:frontier}
\end{figure}

\Cref{fig:frontier} illustrates that many different architectures can participate in the optimal frontier and that we can achieve a smooth variation
between the pure autodecoding models and models that encode more and more semantic and syntactic information.
On the left, we see three syntactic encoders, which do a good job of capturing both the content of the digit and its style, while
having variance in the decodings that seem to capture the sampling noise.
On the right, we have six clear autodecoders, with very low rate and very high variance in the reconstructed or generated digit.
In between are three semantic encoders, capturing the class of each digit, but showing a wide range of decoded style variation,
which corresponds to the syntax of MNIST digits.
Finally, between the syntactic encoders and semantic encoders lies a modeling failure, in which a weak encoder and marginal are
paired with a strong decoder. The rate is sufficiently high for the decoder to reconstruct a good amount of the semantic and syntactic
information, but it appears to have failed to learn to distinguish between the two.

%% file: omniextra.tex
\section{Results on OMNIGLOT}
\label{sec:omniExtra}

\begin{figure*}[hbp]
	\centering
 \subfloat[Distortion vs Rate]{\includegraphics[width=0.47\textwidth]{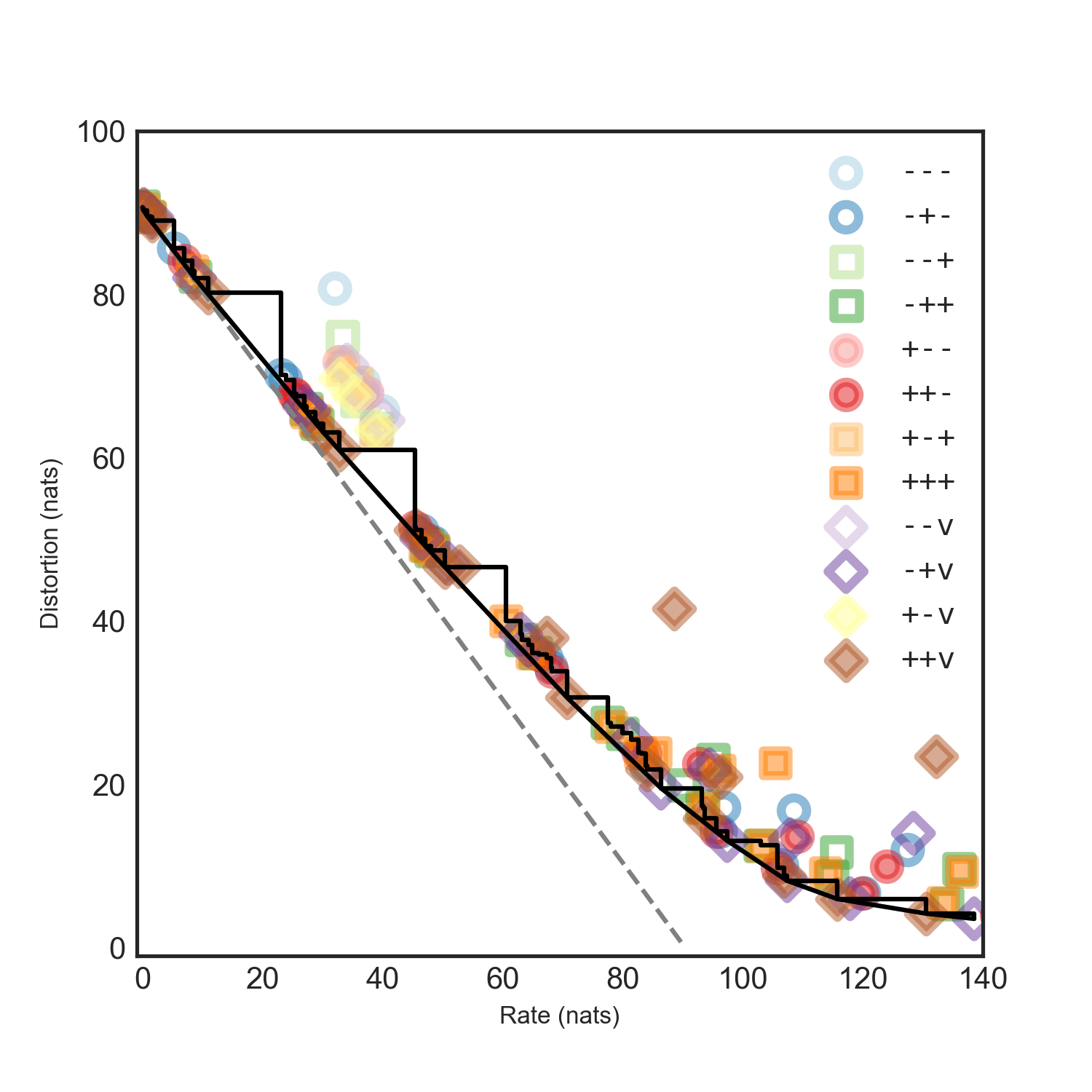}
	\label{fig:omnird}}
	\subfloat[-ELBO ($R+D$) vs Rate]{\includegraphics[width=0.47\textwidth]{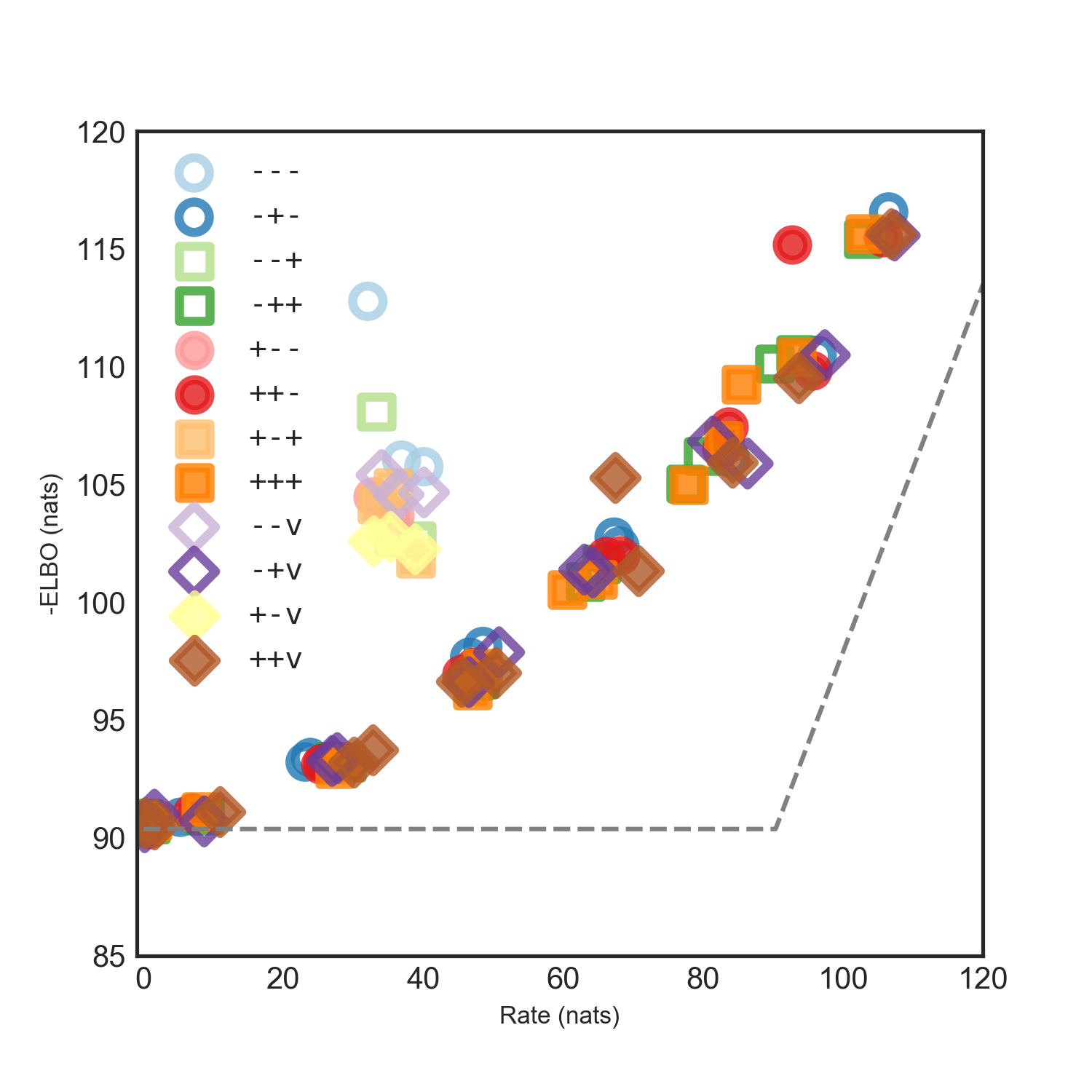}
	\label{fig:omnielbor}} \\
	\subfloat[Distortion vs Rate Breakout]{\includegraphics[width=0.94\textwidth]{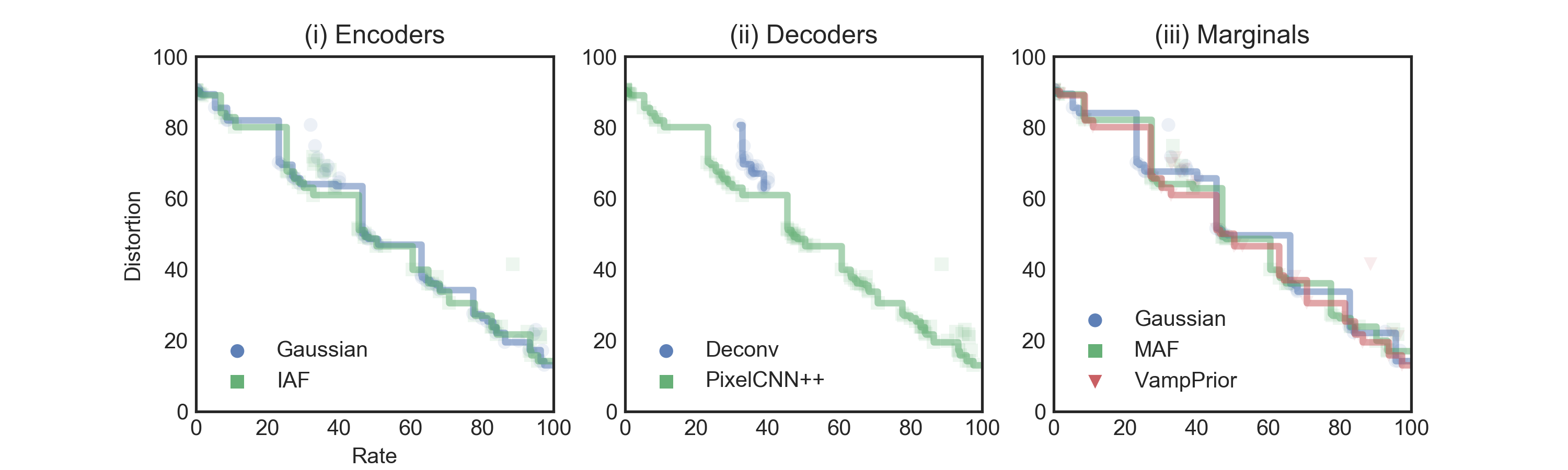}
	\label{fig:omnibreakout}}
  \caption{
    \small
	Results on Omniglot. Otherwise same description as \Cref{fig:mnistRD}.
  (a)  Rate-distortion curves.
  (b) The same data, but on the skew axes of
  -ELBO = $R+D$ versus $R$.
  (c) Rate-distortion curves by encoder, decoder, and marginal family.
  }
  \label{fig:omniRD}
\end{figure*}

\Cref{fig:omniRD} plots the RD curve for various
models fit to
the Omniglot dataset~\citep{lake2015human}, in the same form as the MNIST
results in \Cref{fig:mnistRD}.
Here we explored $\beta$s for the powerful decoder models ranging from 1.1 to 0.1,
and $\beta$s of 0.9, 1.0, and 1.1 for the weaker decoder models.

 On Omniglot, the powerful decoder models dominate over the weaker decoder models.
 The powerful decoder
 models with their autoregressive form most naturally sit at very low rates.  We
 were able to obtain finite rates by means of KL annealing.
 Our best achieved ELBO
 was at -90.37 nats, set by the ++- model with $\beta=1.0$ and KL annealing. This
 model obtains $R=0.77, D=89.60, ELBO=-90.37$ and is nearly auto-decoding.  We
 found 14 models with ELBOs below 91.2 nats ranging in rates from 0.0074 nats to
 10.92 nats.

 \begin{figure*}[tbp]
 	\centering
   \hspace{-.29in}
   \subfloat[Omniglot
   Reconstructions: $z \sim e(z|x)$, $\hat{x} \sim d(x|z)$]{\includegraphics[height=2in]{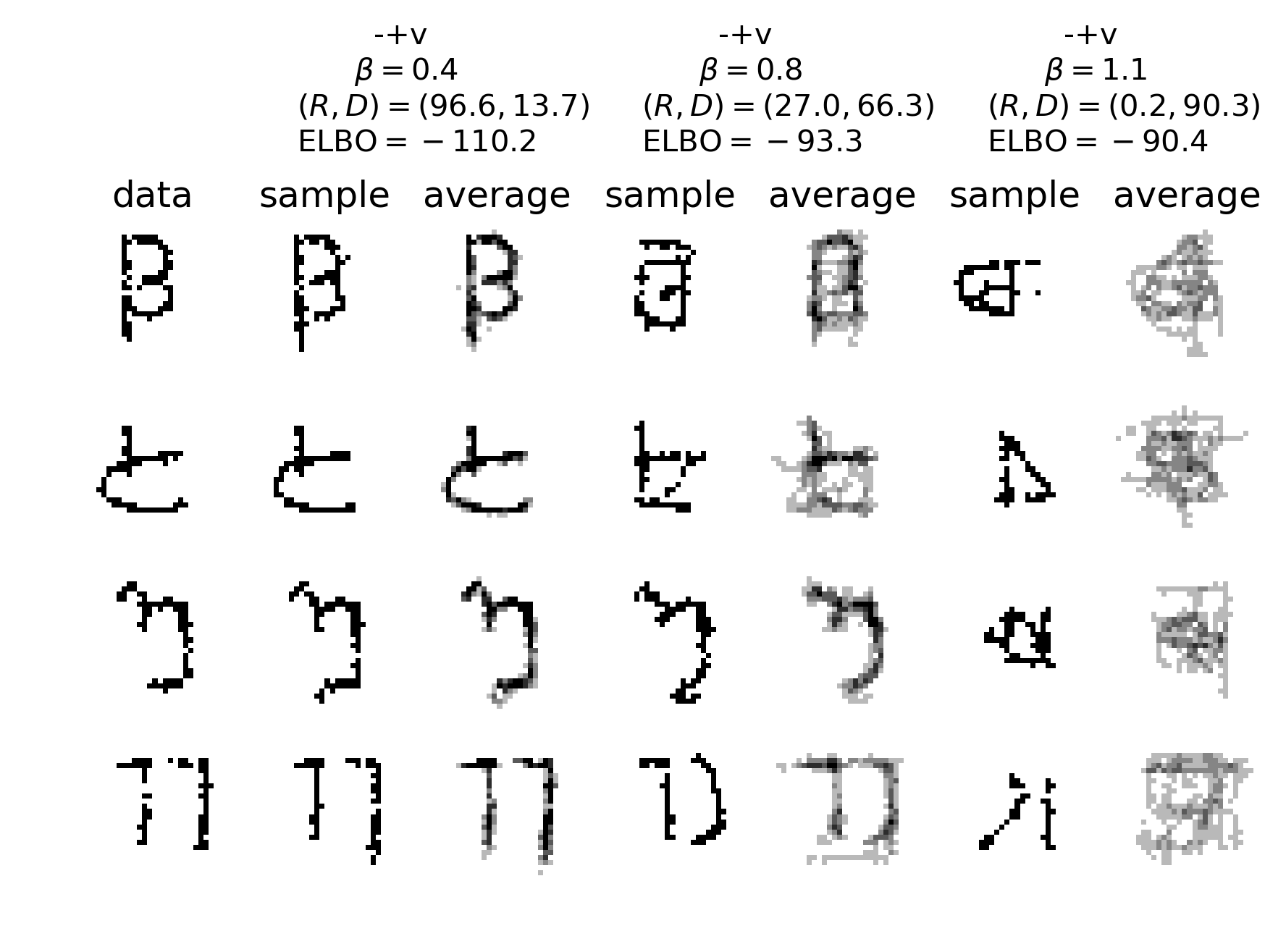}\quad
 	\label{fig:omnirecons}}
   \subfloat[Omniglot Generations: $z \sim m(z)$, $\hat{x} \sim d(x|z)$]{\includegraphics[height=2in]{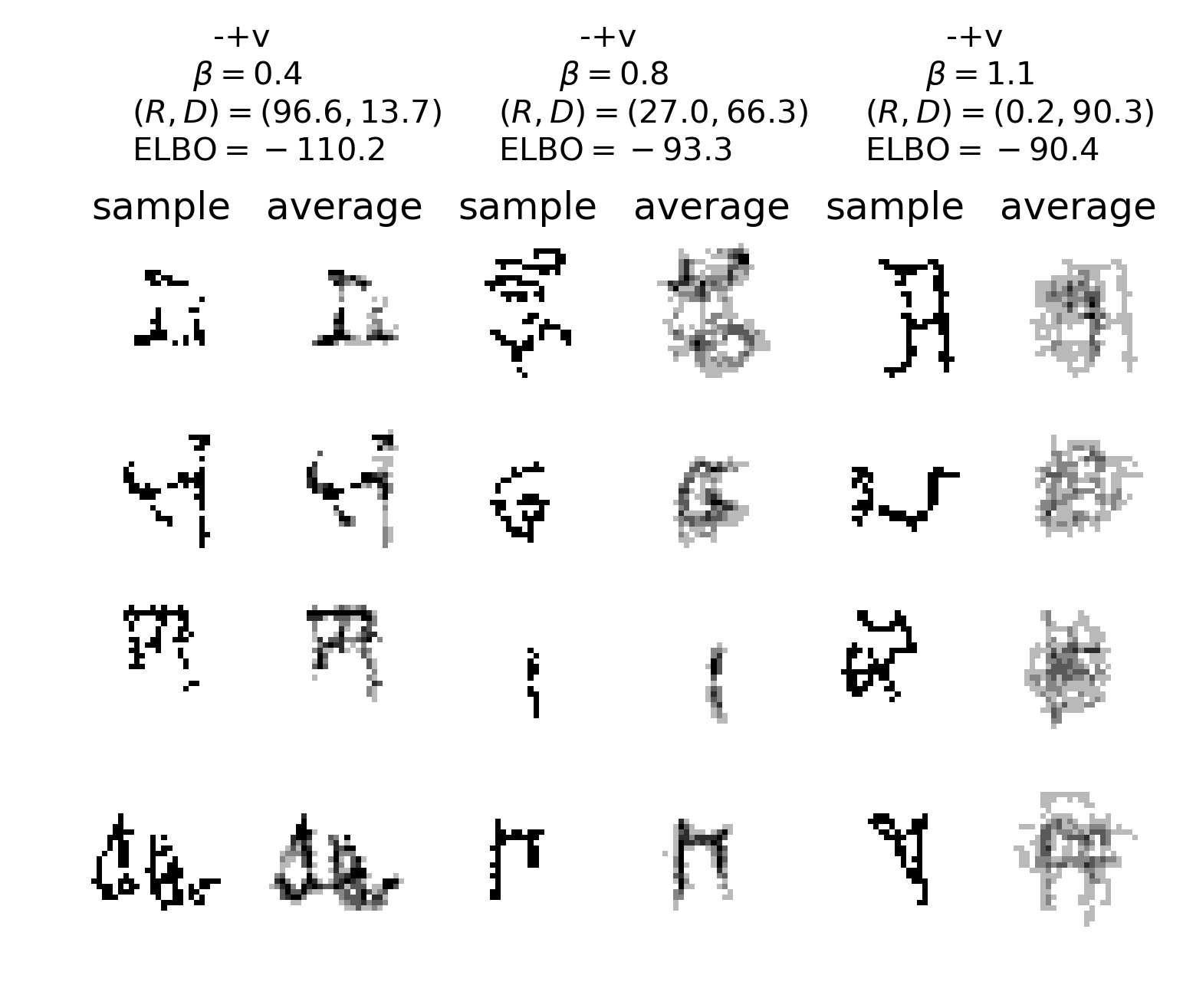}
 	\label{fig:omnigen}}
   \caption{
   \small
 We can smoothly move between pure autodecoding
 	and autoencoding behavior in a single model family by tuning
 	$\beta$.
         (a)  Sampled
 	reconstructions from the -+v model family trained at given $\beta$
 	values. Pairs of columns show a single reconstruction and the mean of 5 reconstructions.
   The first column shows the input samples.
   (b) Generated images from the same set of models. The pairs of columns are single samples and the mean of 5 samples.
   See text for discussion.
   }
   \label{fig:omniSamples}
 \end{figure*}

 Similar to \Cref{fig:mnistSamples} in \Cref{fig:omniSamples} we show sample reconstruction and generated images from the
 same "-+v" model family trained with KL annealing but at various $\beta$s.  Just like in the MNIST case, this demonstrates
 that we can smoothly interpolate between auto-decoding and auto-encoding behavior in a single model family, simply
 by adjusting the $\beta$ value.

\begin{figure}[htbp]
	\centering
	\subfloat[(reconstructions)]{\includegraphics[width=1.0\linewidth]{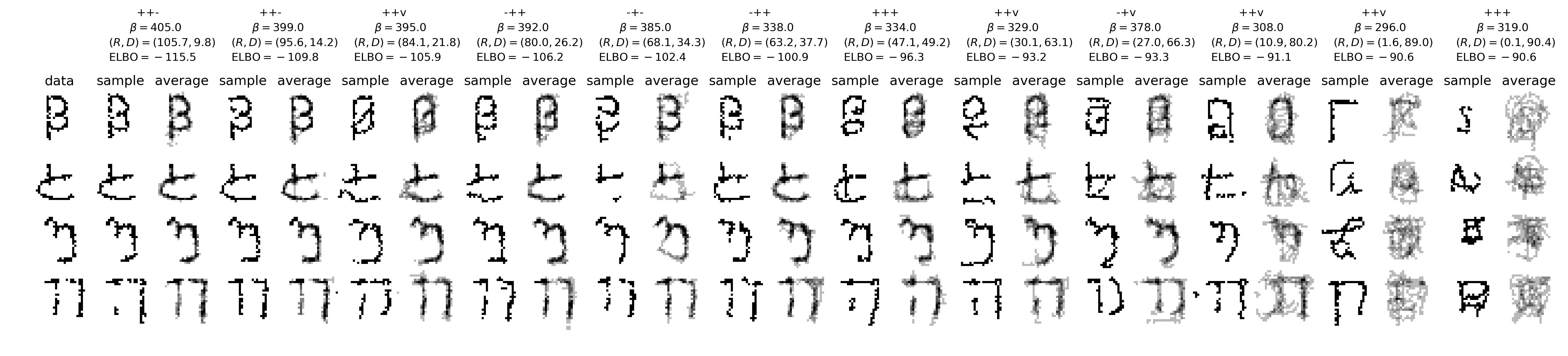}} \\
	\subfloat[(generations)]{\includegraphics[width=1.0\linewidth]{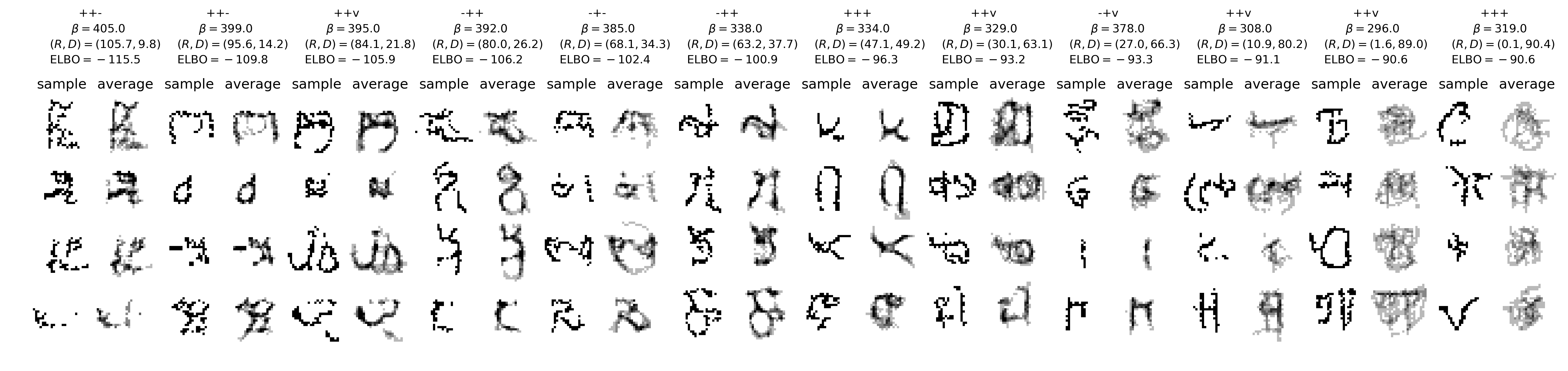}} \\
	\caption{Exploring the Omniglot frontier. Here we show the reconstructions and generated samples from a whole collections of runs that all lie
	on the frontier of relealizable rate distortion tradeoffs.  We do this primarily to illustrate that many different architectures can participate and that we can achieve a smooth variation between the pure generative models and models that encode larger and larger rates.
	}\label{fig:omnifrontier}
\end{figure}

%% file: generative.tex
\section{Generative mutual information}
\label{sec:Igen}

Given any four distributions: $\qp(x)$ -- a density over some data space $X$,
$\qe(z|x)$ -- a stochastic map from that data to a new representational space $Z$,
$\qd(x|z)$ -- a stochastic map in the reverse direction from $Z$ to $X$, and
$\qm(z)$ -- some density in the $Z$ space; we were able to find an inequality
relating three functionals of these densities that must always hold.  We found
this inequality by deriving upper and lower bounds on the mutual information
in the joint density defined by the natural \emph{representational} path through
the four distributions, $\qprep(x,z) = \qp(x)\qe(z|x)$.  Doing so naturally
made us consider the existence of two other distributions $\qd(x|z)$ and $\qm(z)$.
Let's consider the mutual information along this new \emph{generative} path.
\begin{equation}
	\qpgen(x,z) = \qm(z)\qd(x|z)
	\label{eqn:generative}
\end{equation}
\begin{equation}
	\Igen(X;Z) = \iint dx\, dz\, \qpgen(x,z) \log \frac{\qpgen(x,z)}{\qpgen(x)\qpgen(z)}
	\label{eqn:Igen}
\end{equation}

Just as before we can easily establish both a variational lower and upper bound on this
mutual information.  For the lower bound (proved in Section~\ref{sec:lowerboundgen}), we have:
\begin{equation}
	E \equiv \int dz\, p(z) \int dx\, p(x|z) \log \frac{q(z|x)}{p(z)} \leq \Igen
	\label{eqn:E}
\end{equation}
Where we need to make a variational approximation to the decoder posterior,
itself a distribution mapping $X$ to $Z$.  Since we already have such a
distribution from our other considerations, we can certainly use the encoding
distribution $q(z|x)$ for this purpose, and since the bound holds for any
choice it will hold with this choice. We will call this bound $E$ since it gives the
distortion as measured through the \emph{encoder} as it attempts to encode the generated samples
back to their latent representation.

We can also find a variational upper bound on the generative mutual information
(proved in Section~\ref{sec:upperboundgen}):
\begin{equation}
	G \equiv \int dz\, \qm(z) \int dx\, \qd(x|z) \log \frac{\qd(x|z)}{\qq(x)} \geq \Igen
	\label{eqn:G}
\end{equation}
This time we need a variational approximation to the marginal density of our
generative model, which we denote as $\qq(x)$. We call this bound $G$ for the
rate in the \emph{generative} model.

Together these establish both lower and upper bounds on the generative mutual information:
\begin{equation}
	E \leq \Igen \leq G.
	\label{eqn:sandwichgen}
\end{equation}

In our early experiments, it appears as though additionally constraining or targeting values for these
generative mutual information bounds is important to ensure consistency in the underlying 
joint distributions.  In particular, we notice a tendency of models trained with the $\beta$-VAE 
objective to have loose bounds on the generative mutual information when $\beta$ varies away from 1.

\subsection{Rearranging the Representational Lower Bound}

In light of the appearance of a new independent density estimate $\qq(x)$ in
deriving our variational upper bound on the mutual information in the
generative model, let's actually use that to rearrange our variational lower
bound on the representational mutual information.
\begin{equation}
	\int dx\, \qp(x) \int dz\, \qe(z|x) \log \frac{\qe(z|x)}{\qp(x)} =
	\int dx\, \qp(x) \int dz\, \qe(z|x) \log \frac{\qe(z|x)}{\qq(x)} - \int dx\, \qp(x) \log \frac{\qp(x)}{\qq(x)}
	\label{eqn:rearrangedlower}
\end{equation}
Doing this, we can express our lower bound in terms of two reparameterization independent functionals:
\begin{align}
	U &\equiv \int dx\, \qp(x) \int dz\, \qe(z|x) \log \frac{\qd(x|z)}{\qq(x)} \label{eqn:U} \\
	S &\equiv \int dx\, \qp(x) \log \frac{\qp(x)}{\qq(x)} = -\int dx\, \qp(x) \log \qq(x) - H \label{eqn:S}
\end{align}
This new reparameterization couples together the bounds we derived both the representational mutual information
and the generative mutual information, using $\qq(x)$ in both.  The new function $S$ we've described is
intractable on its own, but when split into the data entropy and a cross entropy term, suggests we set a target
cross entropy on our own density estimate $\qq(x)$ with respect to the empirical data distribution that might be
finite in the case of finite data.

Together we have an equivalent way to formulate our original bounds on the representaional mutual information
\begin{equation}
	U - S = H - D \leq I_{\text{rep}} \leq R
	\label{eqn:sandwich2}
\end{equation}

We believe this reparameterization offers and important and potential way to directly control for overfitting.
In particular, given that we compute our objectives using a finite sample from the true data distribution,
it will generically be true that $\KL{\qemp(x)}{\qp(x)} \geq 0$.  In particular, the usual mode we operate in
is one in which we only ever observe each example once in the training set, suggesting that in particular
an estimate for this divergence would be:
\begin{equation}
	\KL{\qemp(x)}{\qp(x)} \sim H(X) - \log N.
\end{equation}

Early experiments suggest this offers a useful target for $S$ in the reparameterized objective that can
prevent overfitting, at least in our toy problems.

%% file: proofs.tex
\section{Proofs}
\label{sec:proofs}

\subsection{Lower Bound on Representational Mutual Information}
\label{sec:lowerbound}

Our lower bound is established by the fact that Kullback-Leibler (KL)
divergences are positive semidefinite

$$ \KL{q(x|z)}{p(x|z)} = \int dx\, q(x|z) \log \frac{q(x|z)}{p(x|z)} \geq 0 $$
which implies for any distribution $p(x|z)$:
$$ \int dx\, q(x|z) \log q(x|z) \geq \int dx\, q(x|z) \log p(x|z) $$

\begin{align*}
	\Irep &= \Irep(X; Z) = \iint dx\, dz\, \qprep(x,z) \log \frac{\qprep(x,z)}{\qp(x)\qprep(z)} \\
	&= \int dz\, \qprep(z) \int dx\, \qprep(x|z) \log \frac{\qprep(x|z)}{\qp(x)} \\
	&= \int dz\, \qprep(z) \left[ \int dx\, \qprep(x|z) \log \qprep(x|z) - \int dx\, \qprep(x|z) \log \qp(x) \right] \\
	&\geq \int dz\, \qprep(z) \left[ \int dx\, \qprep(x|z) \log \qd(x|z) - \int dx\, \qprep(x|z) \log \qp(x) \right] \\
	&= \iint dx\, dz\, \qprep(x,z) \log \frac{\qd(x|z)}{\qp(x)} \\
	&= \int dx\, \qp(x) \int dz\, \qe(z|x) \log \frac{\qd(x|z)}{\qp(x)} \\
	&= \left(-\int dx\, \qp(x) \log \qp(x)\right) - \left(-\int dx\, \qp(x) \int dz\, \qe(z|x) \log \qd(x|z)\right) \\
	&\equiv H - D
\end{align*}

\subsection{Upper Bound on Representational Mutual Information}
\label{sec:upperbound}

The upper bound is established again by the positive semidefinite quality of KL divergence.

$$ \KL{q(z|x)}{p(z)} \geq 0 \implies \int dz\, q(z|x) \log q(z|x) \geq \int dz\, q(z|x) \log p(z) $$

\begin{align*}
	\Irep &= \Irep(X; Z) = \iint dx\, dz\, \qprep(x,z) \log \frac{\qprep(x,z)}{\qp(x)\qprep(z)} \\
	&= \iint dx\, dz\, \qprep(x,z) \log \frac{\qe(z|x)}{\qprep(z)} \\
	&= \iint dx\, dz\, \qprep(x,z) \log \qe(z|x) - \iint dx\, dz\, \qprep(x,z) \log \qprep(z) \\
	&= \iint dx\, dz\, \qprep(x,z) \log \qe(z|x) - \int dz\, \qprep(z) \log \qprep(z) \\
	&\leq \iint dx\, dz\, \qprep(x,z) \log \qe(z|x) - \int dz\, \qprep(z) \log \qm(z) \\
	&= \iint dx\, dz\, \qprep(x,z) \log \qe(z|x) - \iint dx\, dz\, \qprep(x,z) \log \qm(z) \\
	&= \iint dx\, dz\, \qprep(x,z) \log \frac{\qe(z|x)}{\qm(z)} \\
	&= \int dx\, \qp(x) \int dz\, \qe(z|x) \log \frac{\qe(z|x)}{\qm(z)} \equiv R
\end{align*}

\subsection{Optimal Marginal for Fixed Encoder}
\label{sec:optimalmarginal}

Here we establish that the optimal marginal approximation $p(z)$, is precisely
the marginal distribution of the encoder.
$$ R \equiv \int dx\, \qp(x) \int dz\, \qe(z|x) \log \frac{\qe(z|x)}{\qm(z)} $$
Consider the variational derivative of the rate with respect to the marginal
approximation:
$$ \qm(z) \to \qm(z) + \delta \qm(z) \quad \int dz\, \delta \qm(z) = 0 $$

\begin{align*}
	\delta R &= \int dx\, \qp(x) \int dz\, \qe(z|x) \log \frac{\qe(z|x)}{\qm(z) + \delta \qm(z)} - R \\
	&= \int dx\, \qp(x) \int dz\, \qe(z|x) \log \left( 1 + \frac{\delta \qm(z)}{\qm(z)} \right) \\
	&\sim \int dx\, \qp(x) \int dz\, \qe(z|x) \frac{\delta \qm(z)}{\qm(z)}
\end{align*}
Where in the last line we have taken the first order variation, which must
vanish if the total variation is to vanish.  In particular, in order for this
variation to vanish, since we are considering an arbitrary $\delta \qm(z)$, except
for the fact that the integral of this variation must vanish, in order
for the first order variation in the rate to vanish it must be true that for
every value of $x, z$ we have that:
$$ \qm(z) \propto \qp(x) \qe(z|x), $$
which when normalized gives:
$$ \qm(z) = \int dx\, \qp(x) \qe(z|x), $$
or that the marginal approximation is the true encoder marginal.

\subsection{Optimal Decoder for Fixed Encoder}
\label{sec:optimaldecoder}

Next consider the variation in the distortion in terms of the decoding
distribution with a fixed encoding distribution.
$$ \qd(x|z) \to \qd(x|z) + \delta \qd(x|z) \quad \int dx\, \qd(x|z) = 0 $$

\begin{align*}
	\delta D &= -\int dx\, \qp(x) \int dz\, \qe(z|x) \log (\qd(x|z) + \delta \qd(x|z)) - D\\
	&= -\int dx\, \qp(x) \int dz\, \qe(z|x) \log \left( 1 + \frac{\delta \qd(x|z)}{\qd(x|z)} \right) \\
	&\sim -\int dx\, \qp(x) \int dz\, \qe(z|x) \frac{\delta \qd(x|z)}{\qd(x|z)}
\end{align*}

Similar to the section above, we took only the leading variation into account,
which itself must vanish for the full variation to vanish.  Since our variation
in the decoder must integrate to 0, this term will vanish for every $x, z$ we
have that:
$$ \qd(x|z) \propto \qp(x) \qe(z|x), $$
when normalized this gives:
$$ \qd(x|z) = \qe(z|x) \frac{\qp(x)}{\int dx\, \qp(x) \qe(z|x)} $$
which ensures that our decoding distribution is the correct posterior
induced by our data and encoder.

\subsection{Lower bound on Generative Mutual Information}
\label{sec:lowerboundgen}

The lower bound is established as all other bounds have been established, with the positive
semidefiniteness of KL divergences.

$$ \KL{d(z|x)}{q(z|x)} = \int dz\, d(z|x) \log \frac{d(z|x)}{q(z|x)} \geq 0 $$
which implies for any distribution $q(z|x)$:
$$ \int dz\, d(z|x) \log d(z|x) \geq \int dz\, d(z|x) \log q(z|x) $$

\begin{align*}
	I_{\text{gen}} &= I_{\text{gen}}(X; Z) = \iint dx\, dz\, \qpgen(x,z) \log \frac{\qpgen(x,z)}{\qpgen(x)\qpgen(z)} \\
	&= \int dx\, \qpgen(x) \int dz\, \qpgen(z|x) \log \frac{\qpgen(z|x)}{\qm(z)} \\
	&= \int dx\, \qpgen(x) \left[ \int dz\, \qpgen(z|x) \log \qpgen(z|x) - \int dz\, \qpgen(z|x) \log \qm(z) \right] \\
	&\geq \int dx\, \qpgen(x) \left[ \int dz\, \qpgen(z|x) \log \qe(z|x) - \int dz\, \qpgen(z|x) \log \qm(z) \right] \\
	&= \iint dx\, dz\, \qpgen(x,z) \log \frac{\qe(z|x)}{\qm(z)} \\
	&= \int dz\, \qm(z) \int dx\, \qd(x|z) \log \frac{\qe(z|x)}{\qm(z)} \\
	&\equiv E
\end{align*}

\subsection{Upper Bound on Generative Mutual Information}
\label{sec:upperboundgen}

The upper bound is establish again by the positive semidefinite quality of KL divergence.

$$ \KL{p(x|z)}{r(x)} \geq 0 \implies \int dx\, p(x|z) \log p(x|z) \geq \int dx\, p(x|z) \log r(x) $$

\begin{align*}
	I_{\text{gen}} &= I_{\text{gen}}(X; Z) = \iint dx\, dz\, \qpgen(x,z) \log \frac{\qpgen(x,z)}{\qpgen(x)\qm(z)} \\
	&= \iint dx\, dz\, \qpgen(x,z) \log \frac{\qd(x|z)}{\qpgen(x)} \\
	&= \iint dx\, dz\, \qpgen(x,z) \log \qd(x|z) - \iint dx\, dz\, \qpgen(x,z) \log \qpgen(x) \\
	&= \iint dx\, dz\, \qpgen(x,z) \log \qd(x|z) - \int dx\, \qpgen(x) \log \qpgen(x) \\
	&\leq \iint dx\, dz\, \qpgen(x,z) \log \qd(x|z) - \int dx\, \qpgen(x) \log \qq(x) \\
	&= \iint dx\, dz\, \qpgen(x,z) \log \qd(x|z) - \iint dx\, dz\, \qpgen(x,z) \log \qq(x) \\
	&= \iint dx\, dz\, \qpgen(x,z) \log \frac{\qd(x|z)}{\qq(x)} \\
	&= \int dz\, \qm(z) \int dx\, \qd(x|z) \log \frac{\qd(x|z)}{\qq(x)} \equiv G
\end{align*}

%% file: toymodelextra.tex
\section{Toy Model Details}
\label{sec:toymodel-extra}

\paragraph{Data generation.}
The true data generating distribution is as follows.  We first sample a latent
binary variable, $z \sim \Ber(0.7)$, then sample a latent 1d continuous value
from that variable, $h|z \sim \gauss(h|\mu_z,\sigma_z)$, and finally we observe
a discretized value, $x = \mathrm{discretize}(h;\calB)$, where $\calB$ is a set
of 30 equally spaced bins.  We set $\mu_z$ and $\sigma_z$ such that $R^* \equiv \MI(x;z)=0.5$ nats,
in the true generative process, representing the ideal rate target for a latent variable model.

\paragraph{Model details.}
We choose to use a discrete latent representation with $K=30$ values, with an
encoder of the form $\qe(z_i|x_j) \propto -\exp[(w^e_i x_j - b^e_i)^2]$, where
$z$ is the one-hot encoding of the latent categorical variable, and $x$ is the
one-hot encoding of the observed categorical variable.  Thus the encoder has
$2K=60$ parameters.  We use a decoder of the same form, but with different
parameters: $\qd(x_j|z_i) \propto -\exp[(w^d_i x_j - b^d_i)^2]$.  Finally, we
use a variational marginal, $\qm(z_i)=\pi_i$.  Given this, the true joint
distribution has the form $\qprep(x,z) = \qp(x) \qe(z|x)$, with marginal
$\qm(z) = \sum_x \qprep(x,z)$ and conditional $\qprep(x|z) =
\qprep(x,z)/\qprep(z)$.

%% file: mnistdetails.tex
\section{Details for MNIST and Omniglot Experiments}
\label{sec:mnistdetails}

We used the static binary MNIST dataset originally produced for
\citepsupref{binarystaticmnist}\footnote{\url{https://github.com/yburda/iwae/tree/master/datasets/BinaryMNIST}},
and the Omniglot dataset from \citetsupref{lake2015human, iwae}.

As stated in the main text, for our experiments we considered twelve different
model families corresponding to a simple and complex choice for the
encoder and decoder and three different choices for the marginal.

Unless otherwise specified, all layers used a linearly gated activation function
activation function~\citepsupref{gatedactivation}, $h(x) = (W_1 x + b_2) \sigma(W_2 x
+ b_2)$. 

\subsection{Encoder architectures}
For the encoder, the simple encoder was a convolutional encoder outputting parameters to a diagonal Gaussian distribution.
The inputs were first transformed to be between -1 and 1.  The architecture contained 5 convolutional layers, summarized in the format Conv
(depth, kernel size, stride, padding), followed by a linear layer
to read out the mean and a linear layer with softplus nonlinearity to read out the variance of the diagonal Gaussiann distribution.

\begin{itemize}
	\item Input (28, 28, 1)
	\item Conv (32, 5, 1, same)
	\item Conv (32, 5, 2, same)
	\item Conv (64, 5, 1, same)
	\item Conv (64, 5, 2, same)
	\item Conv (256, 7, 1, valid)
	\item Gauss (Linear (64), Softplus (Linear (64)))
\end{itemize}

For the more complicated encoder, the same 5 convolutional layer architecture was used, followed by 4
steps of mean-only Gaussian inverse autoregressive flow, with each step's
location parameters computed using a 3 layer MADE style masked network with 640
units in the hidden layers and ReLU activations.

\subsection{Decoder architectures}
The simple decoder was a transposed convolutional network, with 6 layers of
transposed convolution, denoted as Deconv (depth, kernel size, stride,
padding) followed by a linear convolutional layer parameterizing an independent
Bernoulli distribution over all of the pixels:

\begin{itemize}
	\item Input (1, 1, 64)
	\item Deconv (64, 7, 1, valid)
	\item Deconv (64, 5, 1, same)
	\item Deconv (64, 5, 2, same)
	\item Deconv (32, 5, 1, same)
	\item Deconv (32, 5, 2, same)
	\item Deconv (32, 4, 1, same)
	\item Bernoulli (Linear Conv (1, 5, 1, same))
\end{itemize}

The complicated decoder was a slightly modified PixelCNN++ style
network~\citepsupref{pxpp}\footnote{Original implmentation available at
https://github.com/openai/pixel-cnn}.  However in place of the original RELU
activation functions we used linearly gated activation functions and used six
blocks (with sizes $(28\times 28)$ -- $(14\times 14)$ -- $(7\times 7)$ --
$(7\times 7)$ -- $(14\times 14)$ -- $(28\times 28)$) of two resnet layers in each
block.  All internal layers had a feature depth of 64.  Shortcut connections
were used throughout between matching sized featured maps.  The 64-dimensional
latent representation was sent through a dense lineary gated layer to produce a
784-dimensional representation that was reshaped to $(28\times 28\times 1)$ and
concatenated with the target image to produce a $(28\times 28\times 2)$
dimensional input. The final output (of size $(28\times28\times64)$) was
sent through a $(1\times 1)$ convolution down to depth 1.  These were interpreted
as the logits for a Bernoulli distribution defined on each pixel.

\subsection{Marginal architectures}
We used three different types of marginals. The simplest architecture (denoted (-)), was just a fixed isotropic gaussian distribution in 64 dimensions with means fixed at 0 and variance fixed at 1.

The complicated marginal (+) was created by transforming the isotropic Gaussian base distribution with 4 layers of mean-only
Gaussian autoregressive flow, with each steps location parameters computed
using a 3 layer MADE style masked network with 640 units in the hidden layers
and relu activations. This network resembles the architecture used in \citetsupref{maf}.

The last choice of marginal was based on VampPrior and denoted with (v), which uses a mixture of the
encoder distributions computed on a set of pseudo-inputs to parameterize the prior \citepsupref{vampprior}.
We add an additional learned set of weights on the mixture distributions that are constrained to sum
to one using a softmax function: $m(z) =\sum_{i=1}^N w_i e(z|\phi_i)$ where $N$ are the number of pseudo-inputs, $w$ are the weights, $e$ is the encoder, and $\phi$ are the pseudo-inputs that have the same dimensionality as the inputs.

\subsection{Optimization}
The models were all trained using the $\beta$-VAE objective~\citepsupref{betavae} at
various values of $\beta$.  No form of explicit regularization was used.  The
models were trained with Adam~\citepsupref{adam} with normalized
gradients~\citepsupref{gradientnormalization} for 200 epochs to get good convergence
on the training set, with a fixed learning rate of $3 \times 10^{-4}$ for the
first 100 epochs and a linearly decreasing learning rate towards 0 at the 200th
epoch.